\documentclass[manuscript,screen]{acmart}

\usepackage{listings}

\AtBeginDocument{%
  \providecommand\BibTeX{{%
    \normalfont B\kern-0.5em{\scshape i\kern-0.25em b}\kern-0.8em\TeX}}}

\usepackage{xspace}
\usepackage{tikz}
\usepackage{colortbl}
\newcommand{\zalig}{ZaligVinder\xspace}
\newcommand{\scs}{\textsc{SCS}\xspace}

\newcommand{\var}{\mathcal{X}}
\newcommand{\dom}{\mathcal{D}}
\newcommand{\con}{\mathcal{C}}
\newcommand{\find}{\mathsf{find}\xspace}
\newcommand{\indexof}{\mathsf{indexOf}\xspace}
\newcommand{\contains}{\mathsf{contains}\xspace}
\newcommand{\replace}{\mathsf{replace}\xspace}
\newcommand{\repall}{\mathsf{replaceAll}\xspace}
\newcommand{\regular}{\mathsf{regular}\xspace}
\newcommand{\grammar}{\mathsf{grammar}\xspace}
\newcommand{\toNum}{\mathsf{toNum}\xspace}
\newcommand{\toStr}{\mathsf{toStr}\xspace}
\newcommand{\gcc}{\mathsf{count}\xspace}
\newcommand{\dpllt}{\text{DPLL}(T)\xspace}
\newcommand{\strfuzz}{StringFuzz\xspace}
\newcommand{\gecodes}{\textsc{Gecode}$+$\mathbb{S}\xspace}

\definecolor{dkgreen}{rgb}{0,.6,0}
\definecolor{dkblue}{rgb}{0,0,.6}
\definecolor{dkyellow}{cmyk}{0,0,.8,.3}

\let\_\textunderscore

\lstdefinelanguage{minizinc}
{
	morekeywords={
		ann, annotation, any, array, assert, bool, constraint, else, endif, enum, float, forall, function,
		if, in, include, int, list, of, op, output, par, predicate, record, set,
		solve, string, test, then, tuple, type, var, where,
		abort, abs, acosh, array_intersect, array_union,
		array1d, array2d, array3d, array4d, array5d, array6d, asin, assert, atan, bool2int, card,
		ceil, combinator, concat, cos, cosh, dom, dom_array, dom_size, dominance,
		fix, exp, floor, index_set, index_set_1of2,
		index_set_2of2, index_set_1of3, index_set_2of3, index_set_3of3, int2float, is_fixed,
		join, lb, lb_array, length, let, ln, log, log2, log10, min, max, pow, product, round, set2array,
		show, show_int, show_float, sin, sinh, sqrt, sum, tan, tanh, trace, ub, and ub_array,
		minisearch, search, while, repeat, next, commit, print, post, sol, scope, time_limit, break, fail
	},
	sensitive=false, % are the keywords case sensitive
	morecomment=[l][\em\color{ForestGreen}]{\%},
	morestring=[b]",
}

\begin{document}

%%
%% The "title" command has an optional parameter,
%% allowing the author to define a "short title" to be used in page headers.
\title{A Survey on String Constraint Solving}

%%
%% The "author" command and its associated commands are used to define
%% the authors and their affiliations.
%% Of note is the shared affiliation of the first two authors, and the
%% "authornote" and "authornotemark" commands
%% used to denote shared contribution to the research.
\author{Roberto Amadini}
\email{roberto.amadini@unibo.it}
\affiliation{%
  \institution{University of Bologna}
  \city{Bologna}
  \state{Italy}
}

%%
%% The abstract is a short summary of the work to be presented in the
%% article.
\begin{abstract}
String constraint solving refers to solving combinatorial problems involving constraints over string variables. 
String solving approaches have become popular over the last years given the massive use of strings in different application domains like formal analysis, automated testing, database query processing, and cybersecurity.

This paper reports a comprehensive survey on string constraint solving
by exploring the large number of approaches that have been proposed 
over the last decades to solve string constraints.
\end{abstract}

%%
%% The code below is generated by the tool at http://dl.acm.org/ccs.cfm.
%% Please copy and paste the code instead of the example below.
%%

%%
%% Keywords. The author(s) should pick words that accurately describe
%% the work being presented. Separate the keywords with commas.

%%
%% This command processes the author and affiliation and title
%% information and builds the first part of the formatted document.
\maketitle
\section{Introduction}

Strings are everywhere across and beyond computer science.
They are a fundamental datatype in all the modern programming languages, 
and operations over strings frequently occur in disparate fields such as software analysis, model checking, database applications, web security, bioinformatics and so on~\cite{path_feas,dyn_test_db,mod_check,DBLP:conf/cav/AbdullaACHRRS14,waptec,js-string,aratha,str_mod_check,DBLP:journals/constraints/BarahonaK08}.
Some of the most common uses of strings in modern software development include input sanitization and validation, query generations for databases, automatic generation of code and data, dynamic class loading and method invocation~\cite{str_analysis}.

Reasoning over strings requires handling arbitrarily complex string operations, i.e., relations defined on a number of string variables. In this paper, we refer to these string operations as \emph{string constraints}.
Typical examples of string constraints are string length, 
(dis-)equality, concatenation, substring, regular expression matching.

With the term ``\emph{string constraint solving}'' (in short, string solving or \scs) we refer to the process of modeling, processing, and solving combinatorial problems involving string constraints.
We may see \scs as a declarative paradigm which falls into the intersection between constraint solving and combinatorics on words: the user states a problem with string variables and constraints, and a suitable \emph{string solver} seeks a solution for that problem.

Although works on the combinatorics of words were already published 
in the 1940s~\cite{DBLP:journals/jsyml/Quine46a},
the dawn of a ``more general'' concept of \scs for handling a variety of different string constraints dates back to the late 1980s in correspondence with the rise of \emph{constraint programming} (CP)~\cite{handbookCP} and \emph{constraint logic programming} (CLP)~\cite{clp} paradigms. Pioneers in this field were for example 
Trilogy~\cite{trilogy}, a language providing strings, integer and real constraints, and 
$\text{CLP}(\Sigma^*)$~\cite{clpstar}, 
an instance of the CLP scheme representing strings as regular sets.
%The latter in particular was the first known attempt to use string constraints like regular membership to denote regular sets.

Later in the 1990s and 2000s, string solving has sparked some interest (e.g., \cite{genlang,cpstr,cpgram,hampi09,mona,lpstr,jsa,dprle,php-str}) without however leaving a mark. It was only from the 2010s that \scs finally took hold 
in application domains where string processing plays a central role such as test-case generation, software verification, model checking and web security. 
In particular, a number of proposals based on the \emph{satisfiability modulo theory} (SMT)~\cite{smt} paradigm emerged.
The increased interest in the \scs field is nowadays witnessed, e.g., by the introduction of a string track starting from the 2018 SMT competition~\cite{smtcomp}, the annual challenge for SMT solvers. In 2019, the first workshop on string constraints and applications (MOSCA) has been organized~\cite{mosca}. In 2020, the SMT-LIB standard~\cite{smtlib} officially introduced a theory of Unicode strings 
and string benchmarks are flourishing~\cite{strfuzz,banditfuzz}.

A plausible reason for the growing interest in string solving might be the remarkable performance improvements that constraint solvers have achieved over the last years, both on the CP and the SMT side.
It is therefore unsurprising that nowadays CP and SMT are the main paradigms for solving general \scs problems.
Arguably, the widespread interest in \emph{cybersecurity} has given strong impulse to \scs because strings are often the silent enabler of software vulnerabilities, and a bad string manipulation can
have disastrous effects especially for web applications developed in languages like PHP or JavaScript. 

For example, let us have a look at Listing~\ref{lst:ex} showing a %slightly modified 
snippet of PHP code taken from \cite{str_analysis} and representing a simplified version of a sanitization program taken from 
the web application \textit{MyEasyMarket}~\cite{DBLP:conf/sp/BalzarottiCFJKKV08}.
\begin{figure*}[t]
\begin{lstlisting}[language=php,escapechar=|,label=lst:ex,caption=PHP sanitization code snippet.]
<?php
  $www = $_GET["www"]; |\label{ex:get}|
  $l_otherinfo = "URL";
  $www = preg_replace("/[^A-Za-z0-9 .-@:\/]/", "", $www); |\label{ex:repl}|
  echo $l_otherinfo . ": " . $www;
?>
\end{lstlisting}
\end{figure*}
The program first assigns the query string provided by the user via  
the \texttt{\$\_GET} array to the \texttt{\$www} variable (line \ref{ex:get}). Then, after assigning the 
\texttt{"URL"} string to the \texttt{\$l\_otherinfo} variable, in line \ref{ex:repl} the function \texttt{preg\_replace} is used 
to find all the substrings of \texttt{\$www} matching the pattern 
\texttt{[\^{}A-Za-z0-9 .-@:\textbackslash/]}
and to replace all of them with the empty string.
In this way, all the matched substrings will be deleted from \texttt{\$www}. %after executing \texttt{preg\_replace}. 
After this sanitization, the concatenation of \texttt{\$l\_otherinfo}, \texttt{": "}, and \texttt{\$www} 
is printed.

The code in Listing \ref{lst:ex} contains a subtle bug. Indeed, the goal of the programmer is to delete every character  
that \emph{is not} in the set $\{\texttt{a}, \dots, \texttt{Z}, \texttt{A}, \dots, \texttt{Z}, 0, \dots, 9, \texttt{~},\texttt{.}\texttt{-},\texttt{@},\texttt{:},\texttt{/}\}$. However, the missing escape 
character '\texttt{\textbackslash}' before charater '\texttt{-}' in the matching expression will cause the PHP engine to interpret the pattern "\texttt{.-@}" as the \emph{interval} of characters between '\texttt{.}' and '\texttt{@}' according to the ASCII ordering, i.e., the set $\{\texttt{.}, \texttt{/}, \texttt{0}, \dots, \texttt{9}, \texttt{:}, \texttt{;}, \texttt{<}, \texttt{=}, \texttt{>}, \texttt{?}, \texttt{@}\}$. 
This oversight might lead to dangerous executions because the sanitization fails to remove the symbol \texttt{<}, which can be used to start a potentially malicious script. This is an example of  \textit{cross-site scripting} (XSS) vulnerability.

The automated detection of vulnerabilities like XSS or SQL injections would be very hard without suitable string solving procedures.
For example, as we shall see in Section \ref{sec:tool}, one could model the semantics of the \texttt{preg\_replace} function with a regular expression constraint over \texttt{\$www} and add a ``counterexample constraint'' enforcing the occurrence of '\texttt{<}' in \texttt{\$www}. 
If this constraint is satisfied, the PHP program may not be XSS-safe.

%$\repall(x, q, q')$ constraints replacing all the occurrences of a query string $q$ in a target string $x$ with a fresh string $q'$~\cite{repall}. Note that $\repall$ is not decomposable in general into integer constraints, because we do not know \emph{a priori} how many times $q$ occurs in $x$.
%String solving itself is not trivial in this case because the presence of $\repall$ often implies the \emph{undecidability} of a given theory~\cite{DBLP:journals/pacmpl/ChenCHLW18}.

In this survey we classify the large number of different \scs approaches that have emerged over the last decades into three main categories:
\begin{itemize}
\item[(i)] \emph{Automata-based approaches: }
mainly relying on finite state automata to represent the domain of string variables and to handle string operations.
\item[(ii)]  \emph{Word-based approaches: }
algebraic approaches based on systems of \emph{word equations}.
They mainly use SMT solvers to tackle string constraints.
\item[(iii)]  \emph{Unfolding-based approaches: } 
explicitly reducing each string into a number of contiguous elements denoting its characters. %(e.g., $x$ can be mapped to integer variables or bit-vectors).
\end{itemize}

The goal of this survey is to provide an overview of \scs that can serve as a solid base for both experienced and novice users.
We provide a ``high-level'' perspective of string solving by focusing in particular on its technological and practical aspects, without however ignoring its theoretical foundations.
%The focus of the survey is on the technological aspects. It is therefore out of the scope of this paper the definition of a formal ontology of \scs approaches, or the formalization of a standard language for string constraints.
%As we shall see, CP and SMT are the state-of-the-art technologies for solving \scs problems.

\emph{Paper structure.} 
In Section \ref{sec:scs} we give the necessary background notions. 
In Section \ref{sec:theory} we discuss the theoretical foundations of string solving.
In Section \ref{sec:app} we provide a detailed review of several \scs approaches grouped by the categories listed above. 
In Section \ref{sec:tool} we focus on the technological aspects, by exploring the \scs modeling languages, benchmarks, and applications that have been developed.
In Section \ref{sec:concl} we conclude by giving some possible future directions.

\section{Background}
\label{sec:scs}
In this section we define the basics of string constraint solving.
We recall some preliminary notions about strings and automata theory, and
we define the notions of string variables and constraints.
Then, we show at a high level how \scs problems are handled by CP and SMT paradigms---nowadays the 
state-of-the-art technologies for string solving.

We assume that the reader is familiar with the basic concepts of first-order logic.

\subsection{Preliminaries}
\label{sec:prelim}
Let us fix a finite \emph{alphabet}, i.e., a set 
$\Sigma = \{a_1, \dots, a_n\}$ of $n > 1$ symbols also called 
\emph{characters}. 
A \emph{string} (or a \emph{word}) $w$ is a finite sequence of $|w| \geq 0$ characters 
of $\Sigma$, and $|w|$ denotes the length of $w$ (in this work we do 
not consider infinite-length strings). The empty string is denoted 
with $\epsilon$.
The countable set $\Sigma^*$ of all the strings over $\Sigma$ 
is inductively defined as follows:
\emph{(i)} $\epsilon \in \Sigma^*$;
\emph{(ii)} if $a \in \Sigma$ and $w \in \Sigma^*$, then $wa \in \Sigma^*$.
%\emph{(iii)} nothing else belongs to $\Sigma^*$.
%We use the 1-based array notation to lookup the symbols in a string: 
%$w[i]$ is the $i$-th symbol of string $w$, with $1 \leq i \leq |w|$. 

The string \emph{concatenation} of $v, w \in \Sigma^*$ is denoted by $v \cdot w$ (or simply with $vw$ when not ambiguous). We denote with $w^n$ the \emph{iterated concatenation} of $w$ for $n$ times, i.e., 
$w^0 = \epsilon$ and $w^n = w w^{n-1}$ for $n > 0$.
Analogously, we define the concatenation between \emph{sets of strings}:
 given $V, W \subseteq \Sigma^*$, we denote with 
 $V \cdot W = \{vw \mid v \in V, w \in W \}$ (or simply with $VW$) their 
concatenation and with $W^n$ the iterated concatenation, i.e., 
$W^0 = \{\epsilon\}$ and $W^n = W W^{n-1}$ for $n>0$.
We use the 1-based notation to lookup the symbols in a string: 
$w[i]$ is the $i$-th symbol of string $w$, with $1 \leq i \leq |w|$.
We use the notation $w[i..j]$ as a shortcut for the substring $w[i]w[i+1]\cdots w[j]$; if 
$i > j$, we assume $w[i..j]=\epsilon$.

A set of strings $L \subseteq \Sigma^*$ is called a \emph{formal language}.
Formal languages are potentially infinite sets that can be recognised by models of computation 
having different expressiveness. 
\emph{Finite-state automata} (FSA) are among the best known models of computation for denoting sets of strings.
A FSA is a system $M = \langle \Sigma, Q, \delta, q_0, F \rangle$ with a finite alphabet $\Sigma$, a finite set of states $Q$, a transition function $\delta$ over $Q \times \Sigma$ defining the \emph{state transition} occurring when $M$ is in a state $q \in Q$ and a character $w[i] \in \Sigma$ of an input string $w \in \Sigma^*$ is read for $i=1,\dots,|w|$ (the first character $w[1]$ is always read in the initial state $q_0$). A number $k\geq 0$ of final or accepting states $F$ determine if $w$ belongs to the language $\mathcal{L}(M)$ denoted by $M$ or not. A language recognized by a FSA is called a \emph{regular language}.

Different variants and extensions of FSA have been proposed. 
For example, in a \emph{deterministic} FSA (or DFA) the state transition is always deterministic, i.e., $\delta: Q \times \Sigma \to Q$.
Instead, for a \emph{non-deterministic} FSA (or NFA) the transition function is defined as 
$\delta: Q \times \Sigma \to \mathcal{P}(Q)$ where $\mathcal{P}(Q)$ is the powerset of $Q$.
A $\epsilon$-NFA adds to the NFA formalism the possibility of ``empty moves'' (or $\epsilon$-moves) i.e.,
$\delta: Q \times (\Sigma \cup \{\epsilon\}) \to \mathcal{P}(Q)$.
Other variants are the \textit{Boolean} FSA (or BFA)  
where $\delta: Q \times \Sigma \to \mathcal{B}^Q$ and $\mathcal{B}^Q$ is the set of the $2^{2^{|Q|}}$ Boolean functions over $Q$, and the \textit{alternating} FSA (or AFA), a special case of BFA where the initial 
state is a projection over the states of $Q$.
It is well-known that all the variants mentioned aboved have the same expressiveness, i.e., they all denote regular languages.
Common extensions of FSA are instead the \emph{push-down automata}
(PDA, to recognize \textit{context-free} languages) and the \emph{finite-state transducers} 
(FST, i.e., FSA with input/output tapes defining relations between sets of strings).

\subsection{String variables and constraints}
A \emph{string variable} is a variable that can only be assigned to a string of $\Sigma^*$ or, equivalently, whose \textit{domain} is a formal language of $\Sigma^*$.
%We denote with $\vstr$ the set of all the string variables over $\Sigma$. 
We can classify string variables into three hierarchical classes:
\begin{itemize}
  \item \emph{unbounded-length} variables: they can take any value in $\Sigma^*$
  \item \emph{bounded-length} variables: given an integer $\lambda \geq 0$, they can only take values in 
  $\bigcup_{i=1}^\lambda \Sigma^i = \{w \in \Sigma^* \mid |w| \leq \lambda\}$
  \item \emph{fixed-length} variables: given an integer $\lambda \geq 0$, 
  they can only take values in $\Sigma^\lambda=\{w \in \Sigma^* \mid |w| = \lambda\}$
\end{itemize}

We call \emph{string constraint} a relation over at least a string variable or constant.
To avoid confusion, if not better specified, we will use the uppercase notation for variables and 
the lowercase for constants.
%For example, the concatenation operation is, in our context, a ternary
%string constraint.\footnote{In other contexts it might come more natural to see the concatenation as 
%a function or an operation rather than a constraint.}
% %$\cdot \subseteq \vstr \times \vstr \times \vstr$. 
%Clearly, instead of writing $(X,Y,Z) \in \cdot$ we use a more convenient 
%functional notation $Z = X\cdot Y$.
In this paper we will only consider  \emph{quantifier-free} constraints involving strings and (possibly) integers, e.g., string length or iterated concatenation.
%, by denoting with $\vint$ and $\cint$ the set of all the integer variables and constraints respectively.
%For example, string length is a binary
%constraint $|\cdot| \subseteq \vstr \times \vint$.
%\fixme{In case remove the definition of $\vstr,\cstr,\vint,\cint$!}
%Because the focus of this work is on strings, we will not consider 
%constraints over $\vint - \vstr$.

Table \ref{tab:cons} provides a high-level overview of the main types of string constraints 
that one can find in the literature.\footnote{The $\gcc$ constraint is also referred as \emph{global cardinality count} (GCC)~\cite{mzn-strings}.}
As we shall see in Section \ref{sec:comp}, a modern \scs approach should be able to properly handle most of them, especially when the domains of the variables involved are finite. If they are infinite, decidability issues may arise as we shall see Section \ref{sec:theory}.

It is noteworthy that some of the constraints in Table \ref{tab:cons} are arbitrarily interchangeable. 
For example, $\replace$ can be rewritten in terms of $\find$, length and concatenation constraints as done in \cite{lexfind}. However, e.g., the same constraint can also be rewritten in terms of $\repall$
and other string constraints:
\[
Y = \replace(X, Q, Q') \iff Y = \repall(X[1\,..\,N+|Q|-1], Q, Q') \cdot X[N+|Q|\,..\,|X|] ~\wedge~ N = \find(Q,X).
\]
In the above formulation we assume the character indexing starting from 1, so $w[1]$ and $w[|w|]$ are respectively the first and last character of a string $w$. This choice is also arbitrary: 
one \scs approach 
may index characters starting from zero---as in the case of SMT-LIB standard~\cite{smtlib}, while another one might prefer a more 
mathematical 1-based indexing notation---as in the case of MiniZinc~\cite{minizinc}.
Furthermore, there is no standard naming convention for string constraints. For instance, 
%$\mathsf{substring}(x,y)$ could be denoted as $\contains(y,x)$.
the $\find(X,Y)$ constraint defined in \cite{lexfind} is also referred 
as $\indexof(Y,X)$.

The semantics of $\toNum$ and $\toStr$ constraints, respectively handling string-number and number-string conversions, are also tricky. According to SMT-LIB specifications (see Section \ref{sec:mod}), $N = \toNum(X)$ returns the non-negative integer in base 10 denoted by $X$, if $X \in \{\texttt{0}, \dots, \texttt{9}\}^*$; otherwise $\toNum(X)=-1$.
The function $\toStr(N)$ instead returns the string representation of $N$, if $N \in \mathbb{N}$; otherwise, $\toStr(N)=\epsilon$.\footnote{
In the SMT-LIB 2.6 specifications, the $\toNum$ and $\toStr$ operations are respectively called  \texttt{str.to\_int} 
and \texttt{int.to\_str}
}
This design choice makes totally sense, but it also prevents the conversion of negative numbers. On the other hand, if negative integers are allowed, what value should $\toNum(X)$ return when $X$ does not denote a valid integer? A possible workaround might be to consider these functions as binary predicates 
 $\toNum(X, N)$ and $\toStr(N, X)$ returning false if $X \not\in \{\epsilon,\texttt{-}\} \cdot \{\texttt{0}, \dots, \texttt{9}\}^*$ or $N \not\in \mathbb{Z}$; otherwise, they have the expected semantics. In this way, 
 $\toNum(X,N) \Leftrightarrow \toStr(N,X)$. This choice might be however too strict in contexts where 
 it is fine converting non-numeric strings to numbers. For example, in JavaScript the conversion of a not numeric string to a number returns the string ``\texttt{NaN}''.
\begin{table}
	\caption{Main string constraints. Variables $X,Y,Z,Q,Q'$ are string variables, while variables $I,J,N$  are integer variables.}
	\label{tab:cons}
	\begin{tabular}{cc}
		\hline
		String constraint & Description \\
		\hline
		$X=Y$, $X \neq Y$ & equality, inequality \\
		$X \prec Y$, $X \preceq Y$, $X \succeq Y$, $X \succ Y$  & lexicographic ordering constraint\\
		$N = |X|$ & string length \\
		$Z=X\cdot Y$, $Y = X^N$ & concatenation, iterated concatenation $N$ times\\
		$Y=X^{-1}$ & string reversal \\
		$Y=X[I..J]$ & substring from index $I$ to index $J$ \\
		$N=\find(X,Y)$ & $N$ is the index of the first occurrence of $X$ in $Y$\\
		$Y=\replace(X, Q, Q')$ & $Y$ is obtained by replacing the first occurrence of $Q$ with $Q'$ in $X$ \\
		$Y=\repall(X, Q, Q')$ & $Y$ is obtained by replacing all the occurrences of $Q$ with $Q'$ in $X$ \\
		$N=\gcc(a, X)$ & $N$ is the number of occurrences of character $a$ in $X$ \\
		$N = \toNum(X)$, $X = \toStr(N)$ & $X$ is the string denoting the number $N$ \\
		$X \in \mathcal{L}(\mathcal{R})$ & membership of $X$ in regular language denoted by $\mathcal{R}$\\
		$X \in \mathcal{L}(\mathcal{G})$ & membership of $X$ in context-free grammar language denoted by $\mathcal{G}$ \\
		\hline
	\end{tabular}
\end{table}

%Also, from the constraints in Table \ref{tab:cons} other string constraints can be derived. For example, the constraint $\mathsf{substring}(x,y)$ 
%imposing that $x$ must be substring of $y$ can be defined 
%as $\find(x,y)>0$. 

Finally, note that restricting to a core language by pruning redundant constraints may be appropriate 
for theoretical purposes but it is often counterproductive for practical use. 
For example, \cite{repall} shows that 
having a dedicated CP propagator for $\replace$ is more effective than rewriting it 
into basic string constraints. Hence, formally defining a ``good'' \scs core language is not a goal of 
this paper because the definition of ``good'' is arbitrary and strongly depends on 
the underlying \scs solving approach and application domain.
Nevertheless, we are interested in exploring which types of string constraints 
are handled by different \scs approaches, as we shall see in Section \ref{sec:comp}.

\subsection{\scs and CP/SMT solving}
The ultimate goal of string constraint solving is to determine whether 
or not a set of string constraints is feasible. 
CP and SMT are probably the two main state-of-the-art technologies for solving \scs problems. However, they tackle the (string) constraints in different ways.

\subsubsection{\scs and CP solving}
From a CP perspective, string constraint solving 
means solving a particular case of \emph{constraint satisfaction problem} (CSP). Formally, a CSP is a triple $\mathcal{P}=(\var,\dom,\con)$ where:
$\var = \{X_1, \dots, X_n\}$ are the \textit{variables};
$\dom = \{D(X_1), \dots, D(X_n)\}$ are the \textit{domains}, where for $i = 1, \dots, n$ each $D(x_i)$ is a set of values that
$x_i$ can take; $\con=\{C_1, \dots, C_m\}$ are the \textit{constraints}, i.e., relations over the variables of $\var$ defining the feasible 
values for the variables.
The goal is to find a \emph{solution} of $\mathcal{P}$, which is an assignment $\sigma: \var \to \bigcup{\dom}$ such 
that $\sigma(X_i) \in D(x_i)$ for $i=1,\dots,n$ and 
$(\sigma(X_{i_1}), \dots, \sigma(X_{i_k})) \in C$ for each constraint $C \in \con$ 
defined over variables $X_{i_1},\dots,X_{i_k} \in \var$.
%The CSP notion can be naturally extended to optimization problems: we just 
%add an \emph{objective function} 
%$\varphi: D(x_1) \times \dots \times D(x_n) \to \mathbb{R}$ that maps each 
%solution to a numerical value to be minimised or maximised.
CP solving combines two main techniques:
\emph{(i) propagation}, which works on individual constraints trying to prune the domains of the variables involved until a fixpoint is reached, and \emph{(ii) branching}, which aims 
to actually find a solution via heuristic search. %(the propagation process is not complete in general).

Fixed an alphabet $\Sigma$ we call a CSP with strings, or $\Sigma$-CSP, a 
CSP $\mathcal{P}=(\var,\dom,\con)$ having $k > 0$ string variables 
$\{W_1, \dots, W_k\} \subseteq \var$ such that $D(W_i) \subseteq \Sigma^*$ for $i=1,\dots,k$, and a number of constraints in $\con$  over such variables. To find a solution, a CP solver can try to compile down a 
$\Sigma$-CSP into a CSP with only integer variables~\cite{mzn-strings}, or it can define specialised string propagators and branchers~\cite{gecode_s,dashed-string,sweep-based}. The latter approach 
has proved to be much more efficient.

For example, consider CSP 
$\mathcal{P}=(\{X,Y,N\}, \{\{{ab}, {bc}, {abcd}\}, \Sigma^*, [1,3]\}, \{N = |X|, X=Y^{-1}\})$
where $X,Y$ are string variables with associated alphabet $\Sigma=\{{a},{b},{c},{d}\}$ and 
$N$ is an integer variable. Propagating $N=|X|$ will exclude the string 
${abcd}$ from $D(X)$ because it has length 4, while the domain of $N$ is 
the interval $[1,3]$. An optimal propagator for $X=Y^{-1}$ would narrow the
domain of $Y$ from $\Sigma^*$ to $\{{ba}, {cb}\}$. Note that 
propagation is a compromise between effectiveness (how many values are pruned) and efficiency (the computational cost of pruning), so sometimes 
it makes sense to settle for efficient but sub-optimal propagators.
Then, $N=|X|$ will narrow the domain of $N$ to singleton $\{2\}$, which actually means assigning the value $2$ to $N$. At this stage, a \emph{fixpoint} is reached, i.e., no more propagation is possible: we have to branch on $\{X,Y\}$ to possibly find a solution. Let us suppose that the variable choice heuristics selects variable $X$ and the value choice  heuristics  
assigns to it the value $ab$; in this case the propagator of $Y=X^{-1}$ 
is able to conclude that $Y={ba}$ so a feasible solution for $\mathcal{P}$ (not the only one) is $\{X={ab},Y={ba},N=2\}$.

Virtually all the CSPs referred in the literature have \textit{finite domains}, i.e., the cardinality of each domain of $\dom$ is bounded. Having finite domains guarantees the decidability of CSPs---which are usually NP-complete problems---by enumeration, but at the same time prevents 
the use of unbounded-length string variables for $\Sigma$-CSPs. 
As we shall see in Section \ref{sec:unfo}, the existing CP approaches for string solving do not handle unbounded-length variables.
In fact, although CP provides the 
$\regular(X,M)$ constraint---stating that $X$ must belong to the language denoted by the FSA $M$---it is also true that the string variable (or the array of integer variables) $X$ must have a fixed~\cite{pesant04-regular} or bounded~\cite{regular} length.
The C(L)P proposals we are aware handling unbounded-length strings via regular sets are \cite{clpstar,cpstr,DBLP:conf/inap/KringsSSDE19}. %These automata-based approaches are however outdated.

\subsubsection{\scs and SMT solving}
\label{sec:smt}
In a nutshell, satisfiability modulo theories generalises the Boolean satisfiability problem to decide whether a formula in first-order logic is satisfiable w.r.t. some \emph{background theory} $T$ that fixes the interpretations of predicates and functions~\cite{smt}.
SMT theories can be arbitrarily enriched and also combined together, even though the latter is 
in general a difficult task.
Over the last decades, several \emph{decision procedures} have been developed to tackle the most disparate theories and sub-theories, including the theory of (non-)linear arithmetic, bit-vectors, floating points, arrays, difference logic, and uninterpreted functions. In particular, 
well-known SMT solvers like, e.g., CVC4~\cite{cvc4-str} and Z3~\cite{z3}  implement the \emph{theory of strings} often in conjunction with related theories, such as linear arithmetic for length constraints and regular expressions.

For example, the quantifier-free theory $T_{SLIA}$ of strings (or \emph{word equations}) and linear arithmetic deals with integers and unbounded-length strings in $\Sigma^*$,
where $\Sigma$ is a given alphabet. Its terms are
string/integer variables/constants, concatenation and length.
The formulas of $T_{SLIA}$ are equalities between strings and linear arithmetic constraints. For example, the formula 
$\phi \equiv X= {ab} \cdot Z \wedge |X|+|Y| \leq 5 \wedge ({abcd} \cdot X = Y \vee |X| > 5)$ where 
${a},{b},{c},{d}\in\Sigma$ and $X,Y,Z$ are string variables
is well-formed for this theory. 
Unfortunately, the decidability of $T_{SLIA}$ is still unknown~\cite{mosca}.

As we shall see in Section \ref{sec:word}, over the last years a growing number of modern SMT solvers has integrated the theory of strings. Most of them are based on the $\dpllt$~\cite{DPLLT} procedure. $\dpllt$ is a general 
framework extending the original DPLL algorithm (tailored for SAT solving) to deal with an arbitrary theory $T$ through the interaction between a SAT solver and a solver specific to $T$. In a nutshell, $\dpllt$ lazily decomposes a SMT problem into a SAT formula, which is handled by a DPLL-based SAT solver which in turn interacts with a theory-specific solver for $T$, whose job is to check the feasibility of the model returned by the SAT solver.

As an example, let us consider the $T_{SLIA}$ theory with the above unsatisfiable formula $\phi$. That  formula is \textit{abstracted} into a Boolean formula $\phi'$, 
obtained by treating the atomic formulas of $\phi$ as Boolean variables, and handled by 
a SAT solver which can return ``unsatisfiable'' or a satisfying assignment (this however does not imply that the overall formula is satisfiable). In the latter case, the constraints corresponding to such assignments are distributed to the different theories. 
For instance,
if the formula $X= {ab} \cdot Z \wedge |X|+|Y| \leq 5 \wedge {abcd} \cdot X = Y$ is 
returned, the constraints $X= {ab} \cdot Z$ and ${abcd} \cdot X = Y$ are delivered to the string solver, while $|X|+|Y| \leq 5$ will be solved by an arithmetic solver. Now, the theory solvers can either find that the constraints are $T_{SLIA}$-satisfiable or return \emph{theory lemmas} to the SAT solver. For example, the string solver might return $\neg(X={ab} \cdot Z) \vee |X|=|Z|+2$ to the SAT solver, which will add the corresponding clause to its knowledge base. The SAT solver will then 
produce a new model or return ``unsatisfiable'', and the 
process will be iteratively repeated until either (un-)satisfiability of $\phi$ is proven or a resource limit is reached. The termination of this procedure is not guaranteed in general, even for decidable theories---often solvers use heuristics to speed-up the search.
\section{\scs Theoretical Foundations}
\label{sec:theory}
String constraint solving lays its theoretical foundations in the theory of automata~\cite{automata,automata_1,automata_2,automata_3,automata_4} and
combinatorics on words~\cite{lothaire1997combinatorics,choffrut1997combinatorics,lothaire2002algebraic,lothaire2005applied,berstel2007origins}.
In particular, the concept of \emph{word equation} is central 
and constitutes the starting block for a number of string constraints.
We can define a word equation as a particular string constraint of the form 
$L = R$ with $L, R \in (\Sigma \cup \mathcal{V})^*$ where $\Sigma$ is 
an alphabet and $\mathcal{V}$ is a set of string variables.  
Algebraically speaking, the structure $(\Sigma^*, \cdot, \epsilon)$ 
is a \emph{free monoid} on $\Sigma$. The \emph{free semigroup} on $\Sigma$ is instead the semigroup $(\Sigma^+, \cdot)$ where $\Sigma^+ = \Sigma^* - \{\epsilon\}$. There is therefore a number of theoretical results for free semigroups that can be applied for string solving~\cite{makanin1977,Va_enin_1983,Durnev1995UndecidabilityOT}.

In 1946, Quine~\cite{DBLP:journals/jsyml/Quine46a} proved that the first-order theory of word equations (i.e., word equations with Boolean connectives and quantification over variables) is \textit{undecidable} by proving its equivalence 
with the first-order theory of arithmetic, which is known to be undecidable. However, in 1977 Makanin~\cite{makanin1977} set a milestone in the history of \scs by proving that the satisfiability problem for the \textit{quantifier-free} theory of word equations is decidable.
More than twenty years later, Plandowski---who refers to Makanin's decision procedure as \textit{``one of the most complicated termination proofs existing in the literature''}~\cite{DBLP:journals/jacm/Plandowski04}---showed that this problem is in PSPACE.
Inspired by the work by Plandowski, Jez~\cite{Jez16} used a technique called \textit{local recompression} for reducing the space consumption to $O(n\log{n})$, where $n$ is the size of the input equation, i.e., the sum of the character count of both sides of the equation.

Makanin's, Plandowski's, Jez's and related techniques can be used to decide the satisfiability of word equations, but their computational complexity is often not suitable for string solvers. Conversely, a main theoretical result used by many 
word-based string solvers (see Section \ref{sec:word}) is the \textit{Levi's lemma}~\cite{levi1944semigroups}, sometimes called \textit{Nielsen's transformation} by analogy with the Nielsen transformation for groups~\cite{Fine1995NielsenTA}, stating that for all strings 
$u, v, x, y \in \Sigma^*$ if $uv = xy$ then there exists a string $w \in \Sigma^*$ such that:
\begin{align*}
uw = x & \qquad \text{if } |u| \leq |x| \\
wv = y & \qquad \text{if } |u| \geq |x|
\end{align*}
The Levi's lemma is the main paradigm underlying the Makanin's method, and it is used by most 
SMT solvers for solving word equations. It is particularly useful for \textit{quadratic} word equations~\cite{qwe}, where each variable may occur at most twice. The decidability in this case can be assessed by incrementally building a \textit{finite} graph of substitutions obtained by repeatedly applying the lemma to reduce the size of the equation. Each node of the graph is a word equation $L=R$ with $L,R \in (\Sigma \cup \mathcal{V})^*$, and edges are possibly added after comparing the prefix of $L$ against the prefix of $R$. 
For example, a node of the form $X{ba}XYZ = Y{a}Z{baa}$ defines three outgoing 
edges: one for the case when $X=Y$, one for the case where $X$ prefix of $Y$, and one for $Y$ prefix of $X$.
In the first case, the next node will be ${ba}XXZ = {a}Z{baa}$; in the other cases, it 
will be respectively ${ba}XXYZ = Y{a}Z{baa}$ and $X{ba}YXYZ = {a}Z{baa}$.
The key property, which ensures the termination of the algorithm, is that for each edge $(L=R, L'=R')$ we have $|L'=R'| \leq |L=R|$, where $|\cdot|$ is the number of variables of the equation.
At the end, the original equation is satisfiable if and only if from its node we can reach a leaf of the form $\epsilon = \epsilon$.
This procedure has a polynomial space complexity but, as proven in \cite{qwe} by reduction from 3-SAT, solving quadratic word equations is in general NP-hard even when a single equation is involved.

Clearly, using word equations only is limiting because this formalism cannot capture the disparate string constraints arising from real-world applications (see Table \ref{tab:cons}), and many languages cannot be expressed using word equations alone.
Because of their widespread use in modern programs, two string 
constraints in particular can be considered as important as word equations: 
the string length and the regular expression membership.

Dealing with string length is tricky because this constraint bridges the strings' and the 
integers' domains. A reasonable approach for handling length constraints is to map them to the 
\textit{Presburger arithmetic}~\cite{presb}, which is known to be decidable. Unfortunately, 
reducing to Presburger arithmetic is not always possible. For example, consider the equation 
$X{ab}Y = Y{ab}X$ with $X,Y$ variables and ${a},{b}$ characters.
The set of pairs of integers:
\[
\{(|\sigma(X)|, |\sigma(Y)|) \mid \text{$\sigma$ is a solution of $X{ab}Y = Y{ab}X$} \}
\]
is not definable in Presburger arithmetic because, as proven in \cite{DBLP:conf/atva/LinM18}, this set actually corresponds to the set of pairs $(n,m) \in \mathbb{N} \times \mathbb{N}$ such that:
\[
n = m \vee (n = 0 \wedge \mathsf{Even}(m)) \vee (m = 0 \wedge \mathsf{Even}(n)) \vee
(n,m > 0 \wedge \mathsf{GCD}(n + 2, m + 2) > 1) 
\]
where \textsf{GCD} is the greatest common divisor, which is not Presburger-definable.

At present, the decidability of the theory of word equations and arithmetic over length functions is still a major \textit{open problem} for string constraint solving. However, decidability results have been proved for \emph{fragments} of this theory. For example, in \cite{DBLP:conf/atva/LinM18} Lin et al. defined a class of quadratic word equations with length and regular constraints for which satisfiability is decidable. Their approach extends the Levi-based construction mentioned above with counters addressing the length of each variable. The decidability of the fragment follows from the decidablity of the existential theory of Presburger arithmetic with divisibility~\cite{presb}. 
%An advantage of restricting to fragments is that considering less expressive languages may imply a more tractable computational complexity for decision procedures~\cite{DBLP:conf/lics/LechnerOW15}.

In \cite{DBLP:conf/hvc/GaneshMSR12} Ganesh et al. defined a \emph{solved form} notion for word equations such that a word equation $\omega$ has a solved form if there is a finite set $S$ of formulae 
logically equivalent to $\omega$ and: \emph{(i)} each formula in $S$ is of the form $X = t$, with $X$ variable and $t$ finite concatenation of constants; \emph{(ii)} each variable in $\omega$ occurs exactly once on the 
left hand side of an equation in $S$ and never on the right hand side of an equation in $S$.
For example, the equations $X\mathtt{a} = \mathtt{a}Y \wedge Y\mathtt{a} = X\mathtt{a}$ can be rewritten in solved 
form as $X = \mathtt{a}^i \wedge Y = \mathtt{a}^i$ with $i \geq 0$.
The solved form guarantees that the implied length constraints are linear inequalities, hence their
satisfiability problem is decidable. Although not all word equations can be rewritten in solved form, the authors claimed that most word equations encountered in practice, within the context of software testing and verification, are either in solved form or can be converted into one.
The authors also constructively show that the satisfiability problem for word equations, string length, and regular expressions is decidable provided that the given word equations have a solved form consisting of regular expressions without unfixed parts.\footnote{A formula 
	with unfixed parts, also called a \textit{unifier}, is a formula with free variables representing an infinite family of solutions~\cite{DBLP:conf/stoc/Plandowski06}.}

\begin{figure}
	\centering
	\includegraphics[width=0.45\linewidth]{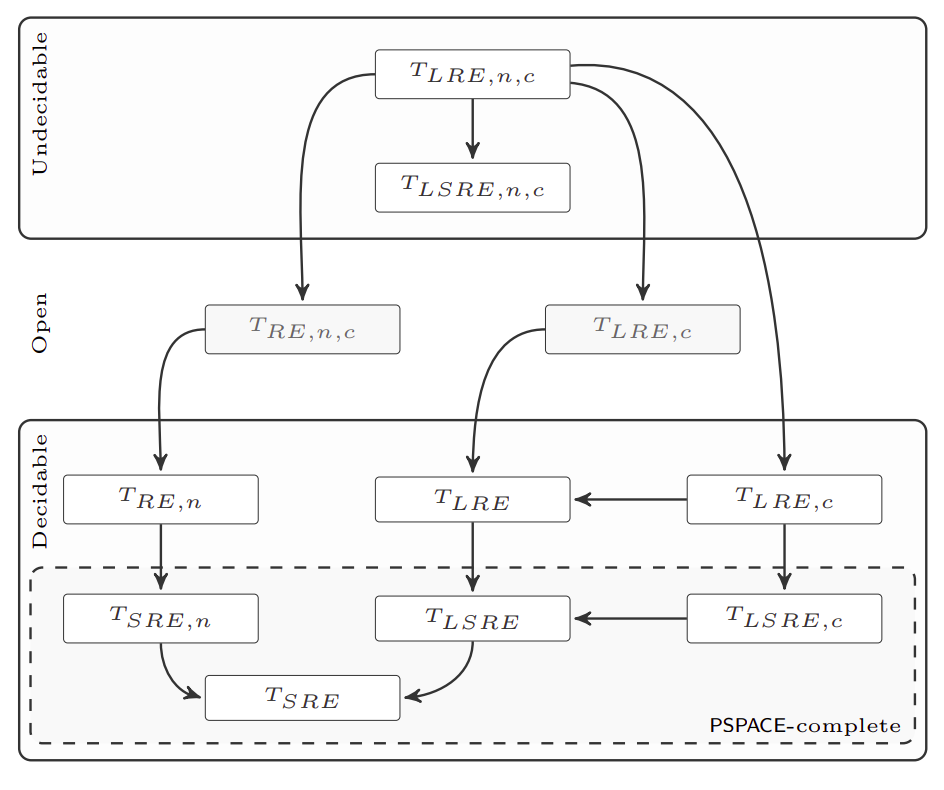}
	\caption{Decidability of various various fragments of $T_{LRE}$. $SRE$ denotes \textit{simple regular expressions}, i.e., regular expressions where the complement operation is not allowed. Arrows model the relation ``is more expressive than''. This picture is taken from \cite{DBLP:journals/corr/abs-2105-07220}.\label{fig:theory}}
\end{figure}

The works in \cite{DBLP:conf/rp/DayGHMN18,DBLP:journals/corr/GaneshB16,DBLP:journals/corr/abs-2105-07220} extend \cite{DBLP:conf/hvc/GaneshMSR12} by considering the (un-)decidability of many fragments of the
first order theory of word equations and their extensions.
For example, the authors prove the undecidability of the theory of word equations with string length, linear arithmetic, and string-number conversion.
In \cite{DBLP:journals/corr/abs-2105-07220} Berzish et al. proved that the quantifier-free first-order theory $T_{LRE,n,c}$ of linear integer arithmetic over string length function ($L$), regex ($RE$) membership predicates,
string-number conversion ($n$), and string concatenation ($c$) in undecidable. 
Note that all of these theories allow equalities between integers but not between strings terms; hence, $T_{LRE,n,c}$ is not able to express general word equations.
The authors also show that several fragments of $T_{LRE,n,c}$ are decidable and some of them are in PSPACE, as summarized in Fig.~\ref{fig:theory}.\footnote{Not all the results in Fig.~\ref{fig:theory} were proven in \cite{DBLP:journals/corr/abs-2105-07220}. For example, the decidability of the theory $T_{LRE,c}$ was already proven in \cite{DBLP:conf/frocos/LiangTRTB15}.}
Interestingly, the decidability of $T_{RE,n,c}$ and $T_{LRE,n}$ is not settled.

In \cite{DBLP:conf/cav/AbdullaACHRRS14}, Abdulla et al. used the notion of \emph{acyclic form} to 
decide the satisfiability of a formula in a fragment of the theory of word equations with arithmetic and regular constraints.
Intuitively, acyclicity is a syntactic property guaranteeing the absence of recursive dependencies
between variables. In general, acyclicity conditions for a formula $\varphi$ can be defined starting from 
the undirected graph $G(\varphi)$ whose nodes are the variables in $\varphi$ and there is an edge $\{X, Y\}$ if there is a constraint in $\varphi$ involving both $X$ and $Y$. The graph $G(\varphi)$ is acyclic if and only if $\varphi$ is in acyclic form. For example, the formula 
$Y = {a}X \wedge Z=X{b} \wedge Y=Z$ 
is not acyclic because the corresponding graph 
$(\{X,Y,Z\}, \{\{X,Y\}, \{X,Z\}, \{Y,Z\}\})$ contains a loop.
In particular, \cite{DBLP:conf/cav/AbdullaACHRRS14} defines an acyclic form preventing variables to appear twice in (dis-)equations. The authors then proved the completeness of their decision procedure for formulae in this acyclic form.

Other interesting string constraints frequently occurring in
problems derived from program analysis are $\replace$ and its generalization $\repall$ (see Table \ref{tab:cons}). These operations are often used, e.g., for sanitizing or transforming a given input string.
In \cite{DBLP:conf/popl/LinB16} the authors show that
any non-trivial theory of strings containing the $\repall$ operation is undecidable unless some kind of \textit{straight-line} restriction is imposed on the formulae.
The undecidability follows by considering $\repall$ constraints as particular cases of finite-state transducers and by reduction from the \textit{Post correspondence problem}~\cite{post1946variant}.
The straight-line fragment is based on straight-line string constraints, which 
intuitively correspond to sequences of assignments in \textit{static single assignment} (SSA) form possibly interleaved with assertions of regular properties. More formally, a relational constraint $\varphi$
is said to be straight-line if it can be rewritten into conjunctions 
$\bigwedge_{i=1}^m X_i = P_i$ such that $X_1, \dots, X_m$ are different variables and each predicate 
$P_i$ contains either variables in $\varphi$ or in $\{X_{1}, \dots, X_{i-1}\}$.
A string constraint is straight-line if it is a conjunction of a straight-line relational 
constraint with a regular constraint.

The straight-line fragment is decidable (the complexity ranges from PSPACE to EXPSPACE) and 
reasonable when handling constraints generated by (dynamic) symbolic execution.
For example, it allows one to express constraints required for analyzing mutated XSS (mXSS) in web applications. Also, this fragment remains decidable in the presence of length, letter-counting, regular, indexOf, and disequality constraints. In \cite{DBLP:journals/pacmpl/ChenCHLW18} Chen et al. also provided a systematic study of the straight-line fragment with $\replace$ and regular membership as basic operations.
\section{\scs Approaches}
\label{sec:app}
In this section we provide a comprehensive overview of the main \scs approaches 
reported in the literature. The classification that we provide into automata-based (Section \ref{sec:auto}), word-based (Section \ref{sec:word}) and unfolding-based (Section \ref{sec:unfo})
approaches follows the categorization provided in \cite{lexfind,ecai-str,dashed-string-jnl}. 
However, similar---if not equivalent---classifications of \scs approaches are reported in many other papers.

For example, back in 2013, \cite{z3str} presented one of the first approaches based on word-equations
(i.e., a word-based approach). The authors stated that:
\textit{``Based on the underlining representation, existing string
analyses can be roughly categorized into two kinds: automata-based [...] and bit-vector-based [...]''}.
We use the term ``unfolding-based'' to generalize the category of bit-vector based approaches. In 
an unfolding-based approach, a string can be unfolded into an ordered sequence of elements having arbitrary type (e.g., integers). The unfolding-based definition allows us to capture \scs approaches like, e.g., PASS~\cite{pass} which models strings with parametric arrays of symbolic length.
The same categorization is repeated two years later in \cite{z3str2}. Importantly, in this paper the term ``word-based'' is introduced for the first time 
to denote \scs approaches handling word equations without abstractions or representation conversions.
In this paper, as also done in \cite{lexfind,ecai-str,dashed-string-jnl}, we stick to this definition. 

Almost at the same time of \cite{z3str}, two other word-based approaches were presented~\cite{cvc4-str,norn}.
In \cite{cvc4-str}, the authors argued that (at that time, 2014) most string solvers were \textit{``standalone tools [...] based on reductions to satisfiability problems over other data types [...] or to automata''}~\cite{cvc4-str}. The approaches based on reductions correspond to what we call unfolding-based problems, to emphasize the ``expansion'' into sequences of elements representing in some way the characters of the string. The same classification of \cite{z3str} is reported in \cite{norn}.

More recently, in \cite{slog}, the authors reported that 
\emph{``string analysis methods are mainly automata-based or satisfiability-based''}. 
Their definition of satisfiability-based method includes both bit-vector and SMT-based approaches.
SMT-based methods are roughly the word-based methods reported in Section \ref{sec:word}.
However, we argue that our distinction is more general because we decouple a given \scs approach from the solving paradigm used to implement it.

We underline that the classification we propose is not a formal ontology, but the result of 
a synthesis among the classifications found in the literature. 
For this reason, as we shall see in Section \ref{sec:hyb}, ``gray areas'' exist corresponding to the hybrid  methodologies integrating different \scs approaches.
Finally, in Sect. \ref{sec:comp} we show how the most recent \scs approaches compare to each other in terms of expressiveness and performance.

\subsection{Automata-based approaches}
\label{sec:auto}
In general, the domain of a string variable is a formal language, i.e., a potentially infinite (yet countable) set. A natural way to denote these sets is through (extensions of) finite-state automata. It is therefore unsurprising that a large number of string solving approaches are based on FSA, possibly enriched with other data structures. %In the following we provide an overview of some of them.

We can say that a \scs approach is automata-based if the solver primarily relies on automata, i.e., string variables are principally represented by automata and string constraints are mainly mapped to corresponding automata operations. As reported in \cite{cvc4-str}, generally speaking there are two sorts of automata-based \scs approaches: the ones where each transition in the automaton represents a single character (e.g., \cite{sushi,stranger}), and the ``symbolic'' ones, where each transition represents a set of characters (e.g., \cite{strsolve, DBLP:conf/wia/Veanes13}).
In the following, we provide a number of approaches---not necessarily string solvers in the strict sense---that may be classified as automata-based. 
%We emphasize that this classification is not meant to be a formal taxonomy, as there are several ``gray areas'' where automata are used together with other solving techniques (see Section \ref{sec:hyb}).

$\text{CLP}(\Sigma^*)$ was one of the first 
attempts to incorporate strings in the CLP framework to strengthen the standard string-handling operations such as concatenation and substring~\cite{clpstar}. This approach was further developed 
by Golden et al.~\cite{cpstr} about 15 years later. Their main contribution is 
to use FSA to represent and manipulate regular sets. The flexibility and expressiveness of using FSA 
however comes at a price: all the operations discussed by the authors (operations on FSA or string constraints like concatenation, containment, length) are linear or quadratic in the size of the FSA representing the string domain.

More recently, in \cite{DBLP:conf/inap/KringsSSDE19} Krings et al. implemented ConString, a CLP system over strings relying on constraint handling rules (CHR)~\cite{chr} in order to generate test data. The domains of the string variables of ConString are represented with FSA as done in \cite{cpstr}. 
ConString is built on top of SWI-Prolog, which does not handle natively FSA. For this reason, 
FSA are encoded as quaternary Prolog terms \texttt{automaton\_dom/4} whose arguments are the set
of states, the transition relation, the set of initial and final states.
However, the authors admit that this representation has several efficiency drawbacks.

In \cite{DBLP:conf/aaai/HansenA07}
Hansen et al. proposed an approach to solve $\Sigma$-CSPs where string constraints are regular expression memberships. The algorithm uses a \textit{Multi}-DFA (MDFA) for joining $n$ different DFAs without reducing to a single DFA, which size would be too big for practical use.
Each MDFA has the form $M = \langle Q, \Sigma, \delta, q_0, \phi \rangle$ where instead of the set of final states $F \subseteq Q$ there is a function $\phi: Q \to \{0,1\}^n$ denoting if a state $q$ is final or not in the $i$-th DFA for $i=1,\dots,n$. 
%For example, suppose having $n=2$ DFAs $M_1,M_2$ and $\phi(q) = (0, 1)$; in this case state $q$ is final in $M_2$ but not in $M_1$.
The authors extend the work of \cite{cpstr} by also using 
\emph{binary decision diagrams} (BDDs~\cite{bdd}) to handle the interactive configuration of variables---not necessarily string variables---having finite domain.
%However, there is no empirical comparison between \cite{DBLP:conf/aaai/HansenA07} and \cite{cpstr}.

The $\regular(A, M)$ global constraint proposed in \cite{pesant04-regular} for solving CP problems 
treats a fixed-size array $A$ of integer variables as a \textit{fixed-length} string belonging to the regular language denoted by a given (non-)deterministic FSA $M$. This constraint was introduced to solve finite-domains CP problems like rostering and car sequencing, and not targeted 
to string solving---in fact, it is a useful constraint 
that has been used in many different CP applications, but its effectiveness diminishes in proportion to the length of $A$. The natural extension of $\regular$ is the $\grammar$ constraint~\cite{cpgram,gramcons}, where instead of a FSA we have a context-free grammar. 
However, despite more expressive, the  $\grammar$ constraint never reached the popularity 
of $\regular$ in the CP community.

An interesting paper about automata-based approaches is \cite{auto-eval}, 
where Hooimeijer et al. study a comprehensive set of algorithms and data
structures for automata operations in order to give a fair comparison between different automata-based \scs frameworks~\cite{mona,dprle,jsa,php-str,rex,stranger,pass}.
According to their experiments, the best results were achieved when using BDDs in combination 
with lazy versions of automata intersection and difference.
BDDs are used to represent UTF-16 character sets: each bit
of a character representation corresponds to a variable of the BDD. 
``Lazy version'' of language intersection and difference 
means that the full automaton is not built: disjointness checking $L(A) \cap L(B) = \emptyset$ and subset checking $L(A) \subseteq L(B)$ are respectively performed instead.

MONA~\cite{mona} is a tool developed in the '90s that acts as a decision procedure for \emph{Monadic Second-Order Logic} (M2L) on finite strings, and as a translator to FSA based on BDDs.
M2L on finite strings, or M2L(Str), joins together (subsets of) \textit{positions} in a string, indexed starting from 0, quantification and Boolean connectives. The language $L(\phi)$ denoted by a M2L(Str) formula $\phi$ is always regular, so each M2L(Str) formula $\phi$ can be translated into an equivalent FSA $M$ such that $L(\phi)=L(M)$.
The key idea of MONA is to \emph{extend} the original alphabet $\Sigma$ with bit vectors for encoding the position information.  The transition function of the resulting FSA is represented via BDDs to overcome the exponential explosion of the extended alphabet.\footnote{As also mentioned in \cite{strsolve}, the BDD representation by MONA is different from that used in \cite{auto-eval}.}
Note that MONA is not a string solver in the strict sense, but the FSA transformation  makes it possible to solve string constraints as a ``side effect''.

FIDO~\cite{fido} is a domain-specific programming formalism built on top of MONA, designed to get a more  high-level and succinct syntax. Its variables can have four different types: \textsf{pos} (positions), \textsf{set} (set of positions), \text{dom} (finite domain) and  \text{tree} ($k$-ary trees). FIDO first compiles the source formula into pure M2L, and then transforms it into a FSA through the MONA tool. 
The FIDO compiler does optimizations at many levels, in most cases relying on the
type structure, to detect simple tautologies and eliminate redundant variables and quantifiers.

Another M2L-based approach is PISA~\cite{pisa}, a path- and index-sensitive tool for static strings 
analysis. PISA can encode a string-manipulating method in \textit{static single assignment} (SSA)
form into an M2L formula $\phi$ representing all the possible strings returned by that method. Then, if $\phi'$ is a M2L formula denoting a set of \textit{unsafe} strings, it checks  $\phi \wedge \phi'$ to verify if the method may return potentially 
dangerous strings. PISA handles index-sensitive operations (e.g., $\indexof$) as well as string replacement operations. It also employs a simple form of path sensitivity, by encoding the branch conditions for a specific variable into M2L predicates. 
PISA was integrated into the taint analysis algorithm used by the IBM Rational AppScan~\cite{appscan}.

JSA~\cite{jsa} is a framework for the static analysis of Java programs. In order to soundly reason about 
string values, it builds a flow graph from Java class files (\textit{front-end}) and then it derives a FSA from such graph (\textit{back-end}). The FSA is derived by first constructing a context-free grammar from the flow graphs with a non-terminal symbol for each node. This grammar is then over-approximated into a regular grammar by using a variant of the \textit{Mohri-Nederhof} algorithm, originally targeted for speech recognition~\cite{Mohri2001}. The resulting regular grammar is finally transformed into a 
a well-founded directed acyclic graph of NFA called \emph{multi-level automata} (MLFA).
The MLFA allows the efficient extraction of a minimal DFA for particular string expressions of interest called \textit{hotspots}.

Inspired by JSA, in \cite{php-str} Minamide developed a string analyzer for the PHP scripting language to detect cross-site software vulnerabilities in a server-side program, and to validate web pages dynamically-generated by the program.
To statically check the generated pages, the string output of a program is approximated via \textit{transducers} by a context-free grammar denoting (a superset of) all the possible output strings possibly generated by the program. Unlike JSA, this tool does not convert the context free grammar into a regular one.

Stranger~\cite{stranger} is another tool for the static analysis of string-related security vulnerabilities in PHP applications. As JSA~\cite{jsa} and the Minamide's tool~\cite{php-str}, Stranger is not a string solver 
in the strict sense but it uses string solving techniques to compute the possible values that string expressions can take during program execution. In particular, Stranger implements an automaton-based approach based on symbolic string analysis. 
It encodes the set of values that string variables can take with a DFA and implements string 
manipulation functions with a symbolic automata-based representation provided by the MONA automata
package. This symbolic encoding also enables Stranger to deal with large alphabets.

Rex~\cite{rex} is a tool based on the Z3 solver~\cite{z3} for symbolically expressing and analyzing regular expression constraints. It relies on 
\emph{symbolic finite-state automata} (SFA) where moves are labeled by 
formulas representing \textit{sets} of characters instead of individual characters.
SFAs are then translated into axioms describing the
acceptance conditions, and Z3 is used for their satisfiability checking and to produce witnesses (i.e., models) for satisfiable formulas.

A more recent string analysis approach is SLOG~\cite{slog}. It is based a on a NFA manipulation engine
with logic circuit representation to support string and automata operations. 
The idea is to avoid determinization as much as possible by performing automata manipulations implicitly 
via logic circuits inspired by those used for industrial applications in electronic design automation.
SLOG also supports symbolic automata and enables the generation of counterexamples. 
The scalability of SLOG is shown for automata with large alphabets in contrast to BDD-based 
representations.

DPRLE~\cite{dprle} is a decision procedure for solving systems of equations over regular language 
variables. The goal here is not to assign satisfying literals to string variables, but instead assigning 
satisfying \emph{regular languages} to the corresponding variables. These variables are involved in constraints of the form $e \subseteq L$ where $L$ is a regular language and $e$ is an expression defined by the concatenation of variables and constant languages.
The authors tackle what they call the \emph{regular matching assignments} (RMA) problem and a subclass, the \emph{concatenation-intersection} (CI) problem. For the RMA problem, DPRLE looks for an assignment that is also \emph{maximal}: for each satisfying variable assignment of the form $X \gets L$, the assignment $X \gets L'$ with $L \subset L'$ is unsatisfiable. The CI problem is a RMA problem allowing the concatenation of at most two variables, each of which has a subset constraint: 
$X \subseteq L \wedge X' \subseteq L' \wedge X \cdot Y \subseteq L''$.
%The authors used DPRLE to find inputs for SQL injection vulnerabilities.

StrSolve~\cite{strsolve} is a decision procedure supporting similar operations to those allowed by DPRLE.
However, StrSolve produces single witnesses rather than atomically generating entire sets of satisfying assignments.
In other words, StrSolve assigns satisfying literals to string variables, rather than assigning regular languages to corresponding variables. Hence the constraints have the form $e \in L$ instead of $e \subseteq L$. This can speed-up the procedure, although the worst-case performance of StrSolve corresponds to that of DPRLE.
The decision procedure of StrSolve is lazy in the sense that string constraints are processed without requiring \textit{a priori} length bounds. This is based on the observation that eager encoding work (i.e., an eager unfolding) is unnecessary if the goal is to find a single solution as quickly as possible.

SUSHI~\cite{sushi} is a string solver based on the 
\emph{Simple Linear String Equation} (SISE) formalism~\cite{sise} to represent path conditions and attack patterns. The application domain is the security analysis of web applications.
A SISE is a string equation of the form $L \equiv R$ where $R$ is a regular expression and 
$L$ a string expression obtainable by concatenation, substring, or substitution of a pattern with 
a string literal. Each string variable can occur at most once in $L$.
SISE uses an automata-based approach to recursively construct ``solution pools'' from which ``concrete solutions'' are derived. As for DPRLE, a concrete solution is a consistent assignment of regular expressions to corresponding variables. To model the semantics of regular substitution, \textit{finite state transducers} (FST) are used.

Luu et al. developed SMC~\cite{smc}, a \emph{model counter} for determining the number of solutions 
of combinatorial problems involving string variables having unbounded length. 
Their work is based on \emph{generating functions}, 
a mathematical tool for reasoning about infinite series that also provides a mechanism to handle the 
cardinality bounds of string sets.
SMC uses FSA to handle conjunctions of regular membership constraints via automata product.
SMC can analyze string operators for C and JavaScript programs.

An Automata-Based model Counter for string constraints (ABC) inspired by SMC is implemented in \cite{abc}.
The ABC solver first constructs an automaton accepting all the feasible solutions of a string constraint, 
and then counts the total number of solutions within a given length bound. 
Model counting corresponds to counting the accepting paths of the resulting automaton up to a given length bound $k$. This is however performed by producing a generating function with the 
\emph{transfer matrix method}~\cite{flajolet2009analytic}. 
In \cite{DBLP:conf/sigsoft/AydinEBBGBY18}  ABC is extended with 
relational and numeric constraints.

An automata-based approach called Qzy is presented in \cite{qzy}, where \textit{Boolean finite automata} (BFA) are used to represent regular expression operations in SMT problems derived from program analysis, in order to avoid the overhead of NFA operations like determinization, complementation and intersection.
The bottleneck here is the emptiness testing: for a NFA is linear, while for 
a BFA is PSPACE-complete. However, in their experiments the authors empirically verified that the IC3 hardware model checking algorithm~\cite{ic3} is effective to decide the BFA emptiness and, by extension, the satisfiability of \scs problems containing regular expressions. 

In \cite{DBLP:journals/jip/ZhuAM19} Zhu et al. proposed a \scs procedure 
where atomic string constraints are represented by 
\emph{streaming string transducers} (SSTs)~\cite{sst}. In particular, the authors use 
bounded SSTs to prove that an input straight-line string constraint is satisfiable if and
only if the domain of the corresponding SST is not empty. This approach can also solve length constraints 
by deriving, and solving via a SMT solver, integer constraints from the \textit{Parikh image} of the SST. This is computed by considering the Parikh vector $\Psi(w) = (c_1, \dots, c_n)$ for each valid output word $w$ of the SST, where $c_i$ is the number of occurrences of the $i$-th symbol of the output alphabet in $w$.

Trau~\cite{trau} is a SMT string solver based on the flattening technique introduced in \cite{flat-trau}, where \emph{flat automata} are used to 
capture simple patterns of the form $w_1^* w_2^* \cdots w_n^*$ where 
$w_1, w_2, \dots, w_n$ are finite words.
Trau relies on a Counter-Example Guided Abstraction Refinement (CEGAR) framework~\cite{cegar} where an under- and an over-approximation module interact to increase the string solving precision.
In addition, Trau implements string transduction by reduction to context-free membership constraints.

Trau has been extended by Z3-Trau~\cite{z3trau} to take advantage of the capabilities of the underlying 
Z3 solver. The main difference between the original Trau and Z3-Trau is that, to handle string-number constraints more efficiently, the latter works fully symbolically with \textit{parametric flat automata} (PFA), i.e., flat automata of the form $(A, \psi)$  where: $A$ is an automaton having an alphabet $V$ of \emph{variables} over an input alphabet $\Sigma \subseteq \mathbb{N}$, 
and $\psi$ is a linear formula over $V$. The parametric words $v_1 \cdots v_k \in V^*$ accepted by a PFA
define its semantics, i.e., the set of all the strings $I(v_1) \cdots I(v_k) \in \Sigma^*$ where $I$ is any interpretation $I: V \to \Sigma \cup \{\epsilon\}$ satisfying $\psi$.

Sloth~\cite{sloth} is based on the reduction of the satisfiability problem for formulae in the straight-line and the acyclic fragment (see Section \ref{sec:theory}) to the emptiness problem for \emph{alternating finite-state automata} (AFA). The emptiness problem can be decided by model checking 
algorithms (e.g., the IC3 algorithm also used by Qzy~\cite{qzy}).
In this way Sloth can handle string constraints with concatenation, finite-state transducers (hence, also $\repall$), and regular constraints. Sloth was the first solver handling complex string constraints derived from HTML5 applications with sanitisation and implicit browser transductions.
Sloth is built on the top of SMT solver Princess~\cite{DBLP:conf/lpar/Rummer08}.

OSTRICH~\cite{ostrich} is a string solver implementing a decision procedure for the
\textit{path feasibility} of \textit{bounded} programs (i.e., programs without with neither loops nor
branching) possibly containing string operations like concatenation, reverse, functional transducers, and $\repall$.
OSTRICH can be seen as an extension of Sloth, with the main difference that OSTRICH also supports variables in both argument positions of $\repall$, while SLOTH only accepts constant strings as the second argument. The empirical evaluation provided by the authors shows that OSTRICH otuperforms Sloth over all 
the benchmarks.
Like Sloth, OSTRICH also extends the SMT solver Princess~\cite{DBLP:conf/lpar/Rummer08}.

It is worth noting that approaches like, e.g., Sloth, OSTRICH or (Z3-)Trau are not fully automata-based but rather lie in the intersection between word-based and automata-based methods (see Section \ref{sec:word}). It would therefore be reasonable to consider them also as word-based methods. We shall further discuss these aspects in Section \ref{sec:hyb}.

\subsubsection*{Pros and Cons}
Automata enable us to represent infinite sets of strings with finite machines, hence they are a natural and elegant way to represent unbounded-length strings. The theory of automata~\cite{automata} is well defined and studied, and non-trivial string solving would be unthinkable without making use of automata results.

Unfortunately, the performance of automata-based approaches for string constraint solving has been hampered by two main factors:
\emph{(i)} the possible state explosion due to automata operations (e.g., the determinization or the intersection of FSA);
\emph{(ii)} the difficulty, in terms of both efficiency and expressiveness, of capturing an exhaustive set of string operations with automata only. Think, e.g., to a simple language such as 
$\{a^n b^n c^n \mid n \in \mathbb{N}\}$ which is not regular and not even context-free.

For these reasons, the effectiveness of ``general'' \scs approaches \emph{purely} relying on automata
is limited: 
modern approaches using automata are in some sense ``hybrid'', i.e., they embed automata within other solving techniques.
However, for some specific applications automata can still be the best choice. For example, 
the abstract interpretation~\cite{CousotC77} of programs containing string expressions 
can take advantage of the expressiveness and the simplicity of automata.

\subsection{Word-based approaches}
\label{sec:word}
We call a \scs approach \emph{word-based} if it \emph{natively} handles theory of word-equations, possibly enriched with other theories (e.g., integers or regular expressions)~\cite{z3str2}. 
Word-based approaches rely on algebraic results (e.g., the Levi's lemma~\cite{levi1944semigroups}) for solving string constraints over the theory of unbounded strings, without systematically reducing to other data structures such as bit vectors or automata. However, this does not mean that bit vectors and automata are not internally used by word-based string solvers.
The natural candidates for implementing these approaches are the SMT solvers, which can incorporate and integrate the theory of strings in their frameworks.

Back in 2014, Liang et al.~\cite{cvc4-str} argued that: \emph{``Despite their power and success as back-end reasoning engines, general multitheory SMT solvers so far have provided minimal or no native support for reasoning over strings [...] until very recently
	the available string solvers were standalone tools that [...] 
	imposed strong restrictions on the expressiveness [...] Traditionally, these solvers were based on reductions to satisfiability problems over other data types''}. However, in the following years the situation has changed, and more and more word-based string solving approaches have emerged.

In \cite{cvc4-str} the authors integrated a word-based \scs approach into the well-known SMT solver CVC4~\cite{cvc4}. They used a $\dpllt$ approach for solving quantifier-free constraints natively over the theory of unbounded strings with length and positive regular language membership. 
The focus was on solving efficiently string constraints arising from verification and security applications.
The authors claimed that CVC4 was the first solver able to reason about a language of mixed constraints including strings together with integers, reals, arrays, and algebraic datatypes. 
Their decision procedure is based on derivation rules repeatedly applied until unsatisfiability is detected or a ``saturated configuration'' is found---in the latter case the input constraints are 
satisfiable. This procedure is sound (when it terminates, the answer is correct) but in general is not refutation-complete (it may not recognize unsatisfiable formulae) and may not terminate due to the unrolling of regular expressions.
The work in \cite{cvc4-str} has been revised and extended in the following years 
to handle further string functions frequently occurring in security and verification applications such as $\contains$, $\indexof$, $\replace$, string to code point conversion~\cite{cvc4-ext1,cvc4-ext2,cvc4-cpoints}.

Meanwhile, almost at the same time, also the well-established SMT solver Z3~\cite{z3}, started to develop string solving capabilities. Z3-str was 
introduced in \cite{z3str} as a \textit{plug-in} of Z3 to solve string 
constraints arising from web application analysis such as length, concatenation, substring, and replace.
Z3-str treats strings as a primitive type, and solves string constraints with a procedure that systematically breaks down constant strings into substrings and variables into subvariables, until the variables are eventually bounded with constant strings or a conflict is detected.
This procedure may not terminate in general, stuck in infinite splitting, when \textit{``overlapping''} string variables occur. This is a well-known problem when solving word equations. 

\begin{figure}
	\centering
	\includegraphics[width=0.4\linewidth]{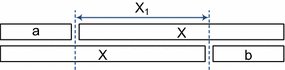}
	\caption{Example of overlapping string variable $X$. Image taken from \cite{z3str2}.\label{fig:overlap}}
\end{figure}

Consider, e.g., the equation $a X = X b$ where $X$ is a string variable and $a,b$ are distinct literals. Intuitively, $X$ is overlapping because if we consider the graphical representation where the substrings of the LHS and RHS of the word equation are aligned as in Fig.~\ref{fig:overlap}, we can observe an overlapping substring of $X$, denoted with $X_1$ in Fig.~\ref{fig:overlap}, that is not equal to $X$ and is ``shared'' between the two sides of the equation. This equation is unsatisfiable because we can rewrite 
$a X = X b$ into an equivalent equation $a a X_1 b = a X_1 b b$, then in turn into $a a a X_2 b b = a a X_2 b b b$, and so on until we eventually reach an unsatisfiable equation of the form $a^{k+1} X_k b^k = a^k X_k b^{k+1}$ with $|X_k| < |a|+|b|$. 
However, if the length of $X$ is unbounded this reasoning cannot be straightforwardly encoded into a terminating procedure.

Z3-str is the progenitor of a number of different string solvers such as Z3str2, Z3strBV, Z3str3, Z3str3RE, and Z3str4.
Z3str2~\cite{z3str2-jnl} extends Z3-str by including overlapping variables detection and new search heuristics.
Overlapping detection allows to individuate cases like that in Fig.~\ref{fig:overlap} to 
avoid common cases of non-termination. However, 
Z3str2 may still not terminate due to the unrolling of regular expressions or the interaction between the integer and string components of its theory. 
The new search heuristics of Z3str2 allows instead the solver to \textit{(i)} prune the search space via bi-directional integration between the string and integer theories, and \textit{(ii)} improve the efficiency of string length queries with a binary search based approach.

Z3strBV aims at the software verification, testing, and security analysis of C/C++ programs.
Instead of mapping strings to bit-vectors, or representing bit-vectors as natural numbers, 
Z3strBV defines a decision procedure combining together the theory of string equations, 
string lengths represented as bit-vectors, and bitvector arithmetic. This procedure 
is similar to the one of Z3str2: the main differences are the reasoning on length 
constraints (Z3strBV uses bit-vectors to model underflow/overflow, while Z3str2 uses integers) and the 
search strategy for length assignments (Z3strBV uses binary search, while Z3str2 uses linear search).

Z3str3~\cite{z3str3} extends Z3str2 by adding a technique called \textit{theory-aware branching} to 
expose the structure of the theory literals to the underlying 
SAT solver.
The Z3 core solver is modified to to take into account 
the structure of the theory literals underlying the Boolean abstraction
of the input formula and prioritize simpler branches over more complex ones. For instance, consider 
the equation $XY=AB$ with $X,Y,A,B$ string variables. Z3str3, as well as Z3str2, handles it by exploring three  possible arrangements: (a) $X=A,Y=B$, (b) $X=AX',X'Y=B$, and (c) $XX''=A,Y=X''B$ where $X',X''$ are fresh 
variables. But in this case, Z3str3 chooses branch (a) because no variable is introduced.

Z3str3RE~\cite{z3str3RE} is an extension of Z3str3 handling regular expressions without reducing 
to word equations. Instead, it uses a length-aware, automata-based approach which is sound and complete for the quantifiers-free theory of regular expressions and linear integer arithmetic over string lengths. The basic idea is to take advantage of implicit and explicit length constraints to prune the search space when using automata-based representations.
Z3str3RE is based on Z3str3 so it also supports string-number conversion, concatenation and other string operations via reduction to word equations. This solver is a clear example of hybrid approach, because it joins together automata-based, word-based and also unfolding-based techniques as we shall see in Section \ref{sec:hyb}.

Z3str4~\cite{z3str4} is, at the moment, the last solver of the Z3str* saga. It differs from previous approaches because it is actually a meta-solver or \emph{portfolio solver}. Indeed, apart from the above mentioned Z3str3, Z3str4 also includes in its portfolio two additional solvers: LAS, a CEGAR-style length abstraction solver, and Z3seq~\cite{z3seq}, a Z3-based solver built on top of Z3 (see Section \ref{sec:unfo}). 
Z3str4 uses algorithm selection~\cite{kotthoff2016algorithm} techniques to select which sequence of algorithms or ``arm'' is supposedly better for a given unforeseen problem instance. The selection is performed according to some static features extracted from that instance.

Another SMT string solver that came around the same time of CVC4 and Z3-str is Norn~\cite{norn}. 
Unlike those solvers, Norn relies on a decision procedure that assumes the \textit{acyclicity} of word equations in order to guarantee its termination (see Section \ref{sec:theory}). 
Starting from a conjunction of (dis-)equalities, membership constraints, and arithmetic constraints Norn  
performs a depth-first search based on rules applied in the following order:
(i) arithmetic constraints checking (ii)  compound disequalities elimination (iii) complex equalities splitting (iv) complex membership constraints splitting (v) satisfiability check of remaining constraints.

S3 is a \textit{symbolic} string solver~\cite{s3} motivated by the analysis of web programs inputs. 
It is called symbolic because it is based on the symbolic representation of string constraints 
derived from the symbolic execution of an input program. 
S3 extends Z3-str~\cite{z3str} to handle regular expressions and other high-level operations such as search, replaceAll, regex match, split, test, exec. In particular, S3 added the definition of recursive functions, 
lazily unfolded during the process of incremental solving. Clearly, this implies that in general the termination of its decision procedure is not guaranteed.

The successor of S3, called S3P~\cite{s3p}, uses a progressive search algorithm that mitigates 
the non-termination issues of recursively defined functions and guides the search towards a ``minimal solution'', i.e., a solution having variables with minimal length. S3P tries to detect if a formula is not progressing towards a target solution by pruning the subtrees rooted in those nodes corresponding to formulas that do non contain the minimal solution of the input formula. 
S3P also implements a conflict clause learning technique to prune the search space. 
The overall procedure of S3 is still not complete, but in practice works better than S3.

The latest version of S3, called S3\#~\cite{s3hash}, implements instead an algorithm for 
counting the models of a formula involving the string constraints handled by S3P.
S3\# uses the reduction rules of S3P but, because a model counter needs to count \emph{all} the solutions, 
it cannot apply directly the same algorithm. Inspired by model counters SMC~\cite{smc} and ABC~\cite{abc} (see Section \ref{sec:auto}), S3\# exhaustively builds a reduction tree in a S3P fashion, but with the difference that each node is now associated with a \textit{generating function} representing its count (i.e., the number of solutions of the corresponding formula).

\subsubsection*{Pros and Cons}

State-of-the-art word-based approaches are more expressive and flexible
than ``pure'' automata-based approaches when it comes to solve generic \scs problems, possibly involving non-regular languages. Moreover, word-based approaches do not directly suffer from the state-explosion issues of automata.
%However, we reiterate that automata are still a fundamental support for a number of string solvers.
Word-based approaches are 
naturally built on the top of well-known SMT solvers and 
can natively handle unbounded-length strings.

One the downside, a fundamental problem of word-based string solvers is that some \scs problems 
are undecidable or still open. For example, at present it is still unclear if the satisfiability problem for the quantifier-free theory of word equations, regular-expression membership predicate and length function is decidable in general. This makes SMT solving inevitably incomplete unless some restrictions over the input language are imposed 
(e.g., things become decidable when considering bounded-length variables).

SMT solvers may also suffer from the performance issues due to the disjunctive 
reasoning of the splitting rule of the underlying \dpllt paradigm~\cite{DPLLT}.
Some experimental evaluations~\cite{dashed-string,sweep-based,lexfind} showed that these solvers may encounter difficulties when dealing with string variables or literals having big length.

\subsection{Unfolding-based approaches}
\label{sec:unfo}
An intuitive way of solving string constraints is to reduce them into 
other well-known data types---for which well-established constraint solving techniques already exist---such as Boolean, integers or bit-vectors.
We call a \scs approach unfolding-based if each string variable is unfolded into an homogeneous sequence of $k\geq 0$ variables of a type $T$, and each string constraint is accordingly mapped into a corresponding operation over $T$.

An unfolding approach inherently needs an \emph{upper bound} $\lambda$
on the string length. So, in general these approaches can handle fixed-length or bounded-length string variables, but cannot deal with unbounded-length variables. One can think to unfolding as a sort of \textit{under-approximation}, where models are bounded by $\lambda$.
Once again, we underline that hybrid approaches exist. For example, automata-based or word-based solvers can internally use unfolding techniques (e.g., by bounding the string length) to boost the search for solutions.
In this scenario, a proper choice of $\lambda$ is clearly crucial.
If $\lambda$ is too small, one cannot capture solutions having not-small-enough string length.
On the other hand, too large a value
for $\lambda$ can significantly worsen the \scs 
performance even for trivial problems.

Another important choice is whether to unfold \emph{eagerly} (i.e., statically, before the actual solving process) or \emph{lazily} (i.e., dynamically, during the solving process). An hybrid approach typically performs the unfolding lazily, by deferring it after applying some higher-level reasoning.

Hampi~\cite{hampi09,hampi12} was probably the first SMT-based approach unfolding
string constraints into constraints over bit-vectors, solved by the 
underlying STP solver~\cite{stp}. Its first version~\cite{hampi09} only allowed one fixed-length string variable. Its subsequent version~\cite{hampi12}
added a number of optimisations, e.g., it provided the support for multiple and bounded-length string variables, word equations and substring extraction.

Kaluza~\cite{kaluza} was the back-end solver used by Kudzu, 
a symbolic execution framework for the JavaScript code analysis. 
Similarly to Hampi, Kaluza dealt with string constraints over bounded-length variables by translating them into bit-vector constraints solved with the STP solver~\cite{stp}. In fact, Kaluza can be 
seen as an extension of the first version of Hampi~\cite{hampi09} to support multiple, bounded-length string variables.

PASS~\cite{pass} is a string solver using parameterized arrays of symbolic length as the main data structure to model strings. The indices and the elements of the array can be symbolic too.
String constraints are converted into quantified expressions handled by an iterative quantifier elimination algorithm generating equisatisfiable un-quantified constraints, that are then solved by 
a SMT solver. Moreover, PASS also uses an automaton to handle regular expressions. So, PASS can be seen as an hybrid approach that combines both unfolding and automata methods.

Mapping strings into bit-vectors is suitable for software analysis applications, especially when it comes to precisely handling overflows via \emph{wrapped} integer arithmetic.
Plenty of SMT solvers for the quantifier-free bit-vector formulas exist (often relying on \emph{bit blasting}, i.e., by converting bit-vector formulae to SAT problems) while the CP support for bit-vectors appears limited~\cite{bit_vector}.

Woorpje~\cite{woorpje} uses a back-end SAT solver to solve bounded-length word equations. 
However, the unfolding performed by Woorpje is not the naive encoding mapping each string position to a fixed number of Boolean variables. 
Woorpje instead first encodes with Boolean variables the automaton denoting 
the solutions of an equation, and then it solves the resulting problem with a SAT solver.
This process is refined by considering the length abstraction of word equations, which allows Woorpje 
to narrow the upper bound of string lengths. The length abstraction is also used to guide the search 
in the automaton.

As an alternative to the Boolean encoding of strings, a straightforward approach is to reason on sequences of characters (or, equivalently, integers). For example, Z3seq~\cite{z3seq} is a Z3-based approach---not belonging to the Z3str* family---supporting a general theory of sequences over arbitrary datatypes.
Hence, Z3seq can be used as a string solver by considering strings as sequences of characters.
Z3seq uses \textit{symbolic Boolean derivatives} to reduce regex constraints without constructing automata. In a nutshell, the symbolic derivative of a regex $R$ is a regex $\delta(R)$ possibly 
enriched with if-then-else conditionals to algebraically manipulate complementation and intersection 
operations. This enable lazy unfolding: the symbolic conditionals directly map to the underlying character theory.

Mapping sting variables to integer variables is straightforward for C(L)P solvers~\cite{mzn-strings}.
For example, in \cite{ocl} the authors describe a lightweight solver relying on 
CLP and \emph{Constraint Handling Rules} (CHR) paradigm~\cite{chr} in order to 
generate large solutions for tractable string constraints in model finding.
This approach unfolds string variables by first labelling their lengths 
and domains, and then their characters.

CP solvers can use fixed-length or bounded-length arrays of integer variables to deal with $\regular$, $\grammar$ and 
other string constraints~\cite{pesant04-regular,cpgram,KadiogluS10,HeFPZ13,Maher09,mzn-strings} without a native support for string variables.
However, as shown in \cite{gecode_s,mzn-strings,dashed-string,sweep-based}, having dedicated \textit{propagators} for string variables can make a difference. As explained in Sect.~\ref{sec:scs}, 
string propagation enables the solver to remove inconsistent values from the domain of string variables 
by performing a proper high-level reasoning for each type of string constraint occurring in the problem.

In \cite{ScottFP13} Scott et al. presented a prototypical
bounded-length approach based on the \emph{affix} domain 
to natively handle string variables.
This domain allows one to reason about the content of string
suffixes even when the length is unknown by using a padding 
symbol at the end of the string. 

The approach in  \cite{ScottFP13}  has been 
subsequently improved~\cite{bound_str,gecode_s,ASTRA:PhD:Scott} with a new structured variable type for strings called \emph{Open-Sequence Representation}, for which 
suitable propagators were defined. If $\lambda$ is the maximum string length, each string variable $X$ is modeled with a pair $(A_X, N_X)$ where $A_X$ is an array of integer variables 
such that $A[i] \in \Sigma \cup \{\epsilon\}$ denotes the $i$-th character of $X$, and $N_X$ is an integer variable denoting the length of $X$. The invariant $A[i]=\epsilon \Leftrightarrow A[j]_{i \leq j\leq \lambda}=\epsilon \Leftrightarrow N_X < i$ is enforced for $i=1,\dots,\lambda$.
This approach has been implemented in the Gecode solver~\cite{gecode} and in \cite{mzn-strings} (and following works) has been referred as $\gecodes$.

To mitigate the dependency on $\lambda$ and enable a lazier unfolding, a different CP approach
based on \emph{dashed strings} has been introduced~\cite{dashed-string-jnl}.
Dashed strings are concatenations of distinct set of strings (called \textit{blocks}) used to represent in a compact way the domain of string variables with potentially very big length.
More formally, given an alphabet $\Sigma$ and a length bound $\lambda$, a dashed string has the form $S_1^{l_1,u_1} \cdots S_n^{l_n,u_n}$ with $0 \leq l_i \leq u_i \leq \lambda$ and $S_i \subseteq \Sigma$ for $i=1,\dots,n$. Each block $S_i^{l_i,u_i}$ denotes the language 
$\gamma(S_i^{l_i,u_i}) = \{w \in S_i^* \mid l_i \leq |w| \leq u_i \}$, and the whole dashed string denotes 
the set of strings 
$\{w \in \gamma(S_1^{l_1,u_1}) \cdot \ldots \cdot \gamma(S_n^{l_n,u_n}) \mid |w| \leq \lambda\}$.

Dashed string propagators have been defined for a number of string constraints (e.g., concatenation, length, find/replace, regular expression membership~\cite{dashed-string,sweep-based,lexfind,regular,repall}) and implemented in the 
G-Strings solver~\cite{g-strings}. These propagators are mainly based on the notion of dashed string \emph{equation}, which can be seen as a semantic unification between dashed strings.
For a detailed description of the dashed string formalism, we refer the interested reader to \cite{dashed-string-jnl}.

\subsubsection*{Pros and cons}
The unfolding-based approaches allows one to take advantage of already defined 
theories and propagators without explicitly implementing support for 
strings. Experimental results show that unfolding approaches, and in particular the CP-based dashed string approach, can be quite effective---especially for satisfiable \scs problems involving long, fixed strings (see, e.g., \cite{dashed-string,sweep-based,lexfind}).

However, unfolding approaches also have a number of limitations. 
The most obvious one is the impossibility of handling unbounded-length 
strings. This can be negligible, provided that a good value of 
$\lambda$ is chosen. Unfortunately, deciding a good value of $\lambda$ may not be always trivial. 
CP solvers may fail on unsatisfiable problems with large domains (because they rely on systematic branching over domain values) and on problems with a lot of logical disjunctions (typically dealt with \emph{reification}). 
For example, as shown in \cite{dashed-string-jnl}, the ``overlapping'' problem $a \cdot X = X \cdot b$ mentioned in Section \ref{sec:word} %, where $X$ is a string variable and $a,b$ are distinct string literals, 
is typically hard for a CP solver because it has to test all the possible values of the domain of $X$ before detecting unsatisfiability. Solving strategies based on \textit{clause learning}~\cite{lcg} would be very helpful in these cases, but unfortunately the learning support for string variables is still an open issue.

\subsection{Hybrid approaches}
\label{sec:hyb}
\begin{figure}[t]
	\centering
	\includegraphics[width=0.6\textwidth]{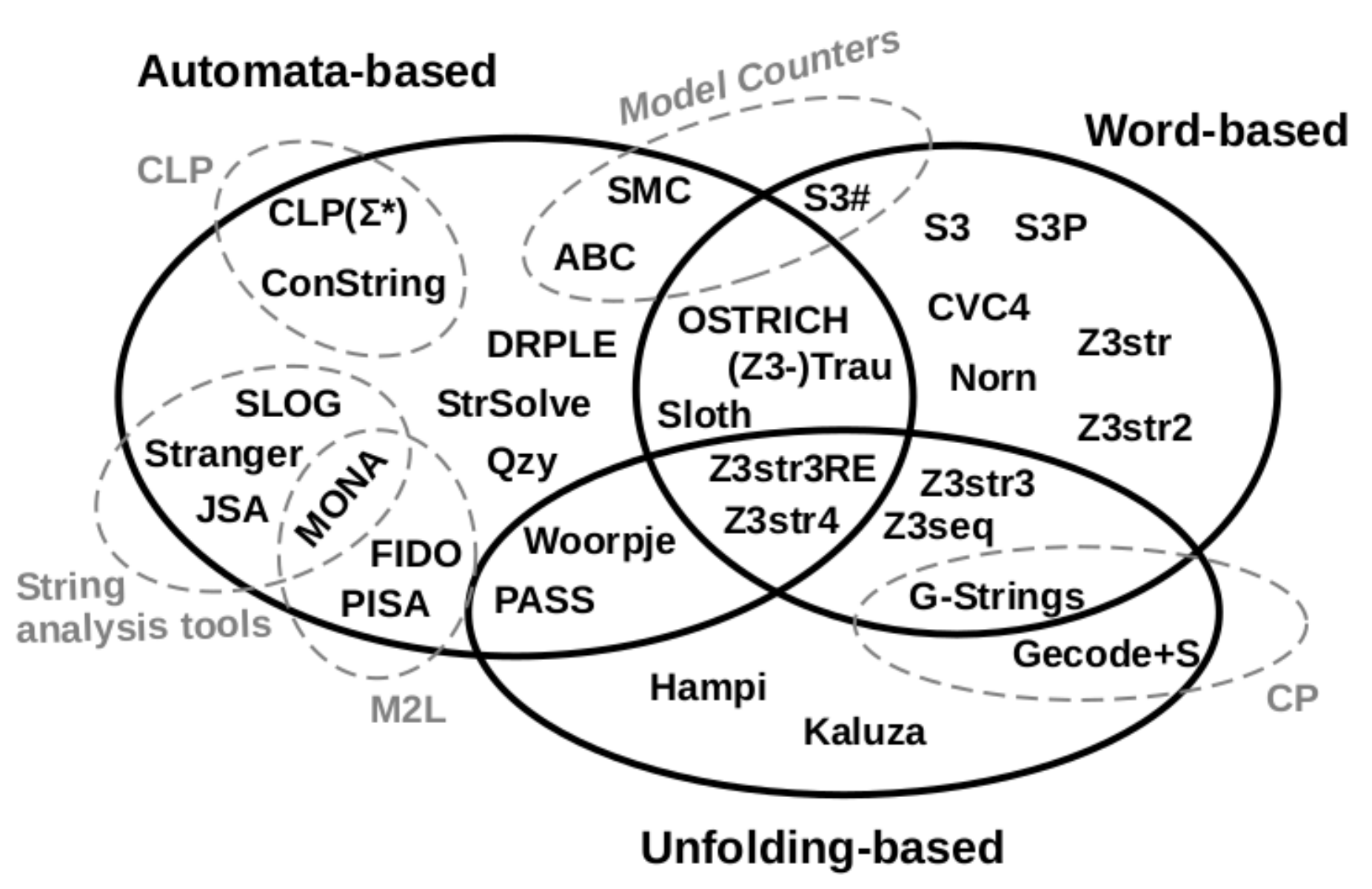}
	\caption{\label{fig:hyb} Graphical representation of \scs approaches.}
\end{figure}
In Section \ref{sec:auto}, \ref{sec:word} and \ref{sec:unfo} we provided an overview of several \scs 
approaches, classified according to the main techniques used to tackle string constraints.
This categorization, as mentioned at the beginning of Section \ref{sec:app}, has been presented---sometimes under different names---in other works. However, the dividing line between these approaches is not so clear-cut. The more string constraints a \scs approach handles, and the more likely is that different \scs techniques are employed to tackle them. This means that some \scs approaches can be referred as \emph{hybrid}, meaning that orthogonal techniques (e.g., automata in combination with unfolding) are combined together to solve string constraints.

The Venn diagram in Fig. \ref{fig:hyb} illustrates the relationships between the \scs approaches mentioned in Sections \ref{sec:auto}--\ref{sec:unfo} by showing, in the overlapping regions of the diagram, the composition of some hybrid approaches.

For example, the PASS approach based on symbolic arrays presented in \cite{pass} is compared against automaton based models (i.e., what we call automata-based approaches) and bit-vector based models (i.e., instances of what we call unfolding-based approaches). The authors make it clear that PASS does not fall in the above categories. Nevertheless, the authors also explicitly say that a contribution of the paper is using interval automata to handle regular expression membership. Hence, we argue that PASS can be considered as a hybrid \scs approach lying at the intersection between automata-based and unfolding-based approaches.

Also the Woorpje solver described in \cite{woorpje} may stand in the middle between automata-based and unfolding-based approaches: it solves word equations by constructing automata that are 
successively encoded into Boolean formulas tackled by a SAT solver.
In \cite{DBLP:conf/icse/DayKMNP20} Woorpje has been extended with a transformation-system approach using CVC4, Z3str3 and Z3seq as assisting string constraint solvers, which are called according to different heuristics.

As mentioned in Section \ref{sec:auto}, in the intersection between automata-based and word-based approaches we can find Sloth, Ostrich, and (Z3-)Trau.
Sloth~\cite{sloth} reduces to alternating finite-state automata to handle string constraints like, e.g., $\repall$. OSTRICH~\cite{ostrich} is an extension of Sloth that uses FSA to compute the pre-image of string constraints or to handle the $\repall$ constraint. Both Sloth and OSTRICH however follow the SMT solving approach and are built on top SMT solver Princess~\cite{DBLP:conf/lpar/Rummer08}.
Analogously, Trau and its extension Z3-Trau can be considered as SMT string solvers, but they also rely on on (parametric) flat automata to handle string constraints.

We put G-Strings~\cite{dashed-string-jnl} halfway between unfolding-based approaches and word-based approaches because, even though it relies on lazy unfolding and it does not actually solve word equations in the strict sense, it is strongly based on the notion of dashed string equation. Roughly, each block $S^{l,u}$ of a dashed string can be seen either as a literal (if the block is fixed, i.e., $l=u$ and $|S|=1$) or a string variable (otherwise). Equating dashed strings does not mean finding a solution, but narrowing the non-fixed blocks by pruning the most inconsistent values.

Z3seq~\cite{z3seq} is a word-based approach that may also be considered unfolding-based because it 
does not implement a theory of strings, but it reduces to the theory of sequences and characters.
We cannot consider it an automata-based approach, because it uses symbolic Boolean derivatives 
to handle regular expressions without actually building automata.

Another solver falling in the word-based/unfolding-based intersection is Z3str3~\cite{z3str3}. In fact, 
in \cite{z3str4} the authors state that: \textit{``The hybrid approach we use in Z3str3 combines the efficiency of an unfolding-based strategy (reduction of fixed-length word equations to bit-vectors) with the ability of a word-based strategy to reason about string terms of unbounded length (the arrangement method).''}

An interesting case is Z3str3RE~\cite{z3str3RE}, an approach that we can consider orthogonal to all classes. Indeed, it is word-based and unfolding-based as explained above. However, it is also automata-based because it uses a length-aware, automata-based approach (see Sect. \ref{sec:word}).

Finally, we can say that Z3str4~\cite{z3str4} is hybrid by definition, being actually a portfolio  joining together different \scs solvers (including Z3str3RE).

\subsection{Comparison}
\label{sec:comp}
\newcommand{\oursolv}{\cite{DBLP:journals/jip/ZhuAM19}\xspace}
\newcommand{\clg}{\cellcolor{lightgray}}

Let us conclude Section \ref{sec:app} by giving a closer look on how the \scs techniques mentioned 
in Sect.\ref{sec:auto}--\ref{sec:unfo} compare to each other.
These approaches, and many others, have been tested on several string benchmarks---especially coming from the software verification world. To give an idea of the comparisons that have been performed between different string solvers, we considered all the approaches listed in Sections \ref{sec:auto}--\ref{sec:unfo} and published from 2015 onward, namely:~\cite{abc,DBLP:conf/sigsoft/AydinEBBGBY18,DBLP:conf/inap/KringsSSDE19,slog,woorpje,DBLP:journals/jip/ZhuAM19,sloth,ostrich,cvc4-ext2,cvc4-ext1,cvc4-cpoints,z3str2-jnl,z3str3,z3strbv,z3str3RE,z3str4,norn,trau,z3trau,s3p,s3hash,ocl,woorpje,gecode_s,bound_str,dashed-string,sweep-based,dashed-string-jnl,lexfind,regular,repall}.
\begin{table}[t]
	\caption{Number of times the approach on that row has been compared against the approach on that column. Cells with number greater than zero have gray background.\label{tab:tools}}
	\scalebox{0.87}{
		\begin{tabular}{c|cccccc|cccc|ccc||c}
			~& Slog & \oursolv & Qzy & Trau* & Sloth & Ostrich & CVC4 & Z3str* & Norn & S3* & Z3seq & G-Strings & Woorpje & \textit{\#dist.} \\
			\hline
			Slog & -- & 0 & 0 & 0 & 0 & 0 & \clg 1 & \clg 1 & \clg 1 & 0 & 0 & 0 & 0 & 3 \\
			\oursolv &  & -- & 0 & 0 & \clg 1 & \clg 1 & 0 & 0 & 0 & 0 & 0 & 0 & 0 & 2 \\
			Qzy & 0 & 0 & -- & 0 & 0 & 0 & 0 & 0 & \clg 1 & 0 & 0 & 0 & 0 & 1 \\
			Trau* & 0 & 0 & 0 & -- & 0 & 0 & \clg 2 & \clg 2 & 0 & \clg 1 & \clg 1 & 0 & 0 & 4 \\
			Sloth & 0 & 0 & 0 & 0 & -- & 0 & \clg 1 & 0 & 0 & \clg 1 & 0 & 0 & 0 & 2 \\
			Ostrich & 0 & 0 & 0 & 0 & \clg 1 & -- & \clg 1 & \clg 1 & 0 & 0 & 0 & 0 & 0 & 3 \\
			\hline
			CVC4 & 0 & 0 & 0 & 0 & 0 & \clg 1 & -- & \clg 1 & 0 & 0 & \clg 3 & 0 & 0 & 3 \\
			Z3str* & 0 & 0 & 0 & \clg 1 & 0 & \clg 1 & \clg 3 & -- & \clg 1 & \clg 2 & \clg 1 & 0 & 0 & \textbf{6} \\
			Norn & 0 & 0 & 0 & 0 & 0 & 0 & \clg 1 & \clg 1 & -- & \clg 1 & 0 & 0 & 0 & 3 \\
			S3* & 0 & 0 & 0 & 0 & 0 & 0 & \clg 1 & \clg 2 & \clg 1 & -- & 0 & 0 & 0 & 3 \\
			\hline
			Z3seq & 0 & 0 & 0 & \clg 1 & 0 & \clg 1 & \clg 1 & \clg 1 & 0 & 0 & -- & 0 & 0 & 4 \\
			G-Strings & 0 & 0 & 0 & 0 & 0 & 0 & \clg 1 & \clg 1 & \clg 1 & 0 & \clg 1 & -- & 0 & 4 \\
			Woorpje & 0 & 0 & 0 & 0 & \clg 1 & 0 & \clg 1 & \clg 1 & \clg 1 & 0 & \clg 1 & 0 & -- & 5 \\
			\hline\hline
			\textit{\#dist.} & 0 & 0 & 0 & 2 & 3 & 4 & \textbf{10} & 9 & 6 & 4 & 5 & 0 & 0 &
	\end{tabular}}
\end{table}
Table \ref{tab:tools} shows how many times, in the papers above, the approach on that row has been compared against the approach on that column.
We excluded from the count the self-comparisons (i.e., comparisons between different variants of the  solver presented in the paper) and the approaches based on model counters, because they only 
compare to each others. Also, we did not report the approaches in \cite{DBLP:conf/inap/KringsSSDE19,ocl,gecode_s}  because they did not compare against any other approach in Table \ref{tab:tools}.
We grouped together ``families'' of solvers: Z3str* indicates all the Z3str-based solvers seen in Section \ref{sec:word}, S3* denotes both S3 and S3P, Trau* refers to both Trau and Z3-Trau. 
The entry \oursolv denotes the unnamed solver described by Zhu et al. in \cite{DBLP:journals/jip/ZhuAM19}, where it is simply referred as ``Our Solver''.
%The entry CP* indicates a CP solver over integers solving a \scs problem by means of the static string-to-integer conversion $\mathcal{F}^{int}$ described in \cite{mzn-strings}.

Each entry on the last row denotes how many \textit{distinct} approaches have been compared against the approach on that column;
each entry on the last column denotes how many distinct approaches the approach on that row has been compared to.
As we can see, most of the comparisons are between SMT-based solvers. Among them, Z3str* and CVC4 solver 
are certainly the most popular. This is not surprising given that Z3 and CVC4 are well-established SMT solvers 
providing a solid base for building \scs capabilities by extension.
A curious case is represented by G-Strings and Woorpje: they are compared against different string solvers, but no paper reports a comparison with them. If for the latter the reason for this asymmetry might be that Woorpje is quite a recent approach, for G-Strings the main reason is likely the absence of a standard interface for modeling \scs problems. On the other hand, as we shall see in Section \ref{sec:mod}, SMT solvers can rely on the SMT-LIB standard.

\subsubsection{Expressiveness}
Let us now focus on the expressiveness of the \scs approaches in Table \ref{tab:tools}.

Slog~\cite{slog} handles operations on automata, including the $\replace$ operation, for acyclic constraints. However, a big limitation of Slog is that it does not support string length operations.
As reported in \cite{ostrich}, also Sloth~\cite{sloth} does not support length constraints. OSTRICH extends Sloth 
by extending the class of transducers it supports. In particular, OSTRICH allows variables in both argument positions of $\repall$, while Sloth only accepts constant strings as the second argument.
According to \oursolv, also the OSTRICH support of integer constraints is limited since \textit{``in
general these constraints do not follow its restrictions''}. However, this issue is not detailed in the 
paper, so it is hard to understand what restrictions they exactly refer to, or what they mean by ``in general''.
Qzy~\cite{qzy} supports Perl-compatible regular expressions by encoding Boolean combination of regular language membership constraints with Boolean finite automata. Besides regular expression membership, there is no mention of other string operations.
Trau~\cite{trau} supports word equations, length constraints, context-free membership queries, and transducer constraints. Z3-Trau extends Trau by also adding the support for string-number constraints.

Norn~\cite{norn} solver supports word equations, length constraints, and regular membership operations but it does not handle constraints like $\indexof$, $\replace$ and $\repall$. These operations are instead 
supported by the solvers of the S3* family~\cite{s3,s3p,s3hash}, which also handle high-level string operations such as search, match, split, test, exec. Woorpje~\cite{woorpje} instead currently only supports word equations with linear length constraints.

CVC4 and Z3*-based are mature solvers that nowadays support plenty of string constraints. They started with word equations, string length and regular expression memberships, and incrementally added the support for new constraints 
such as, e.g., $\indexof$, $\replace$, and the lexicographic ordering.\footnote{More details here: \url{http://cvc4.cs.stanford.edu/wiki/Strings} and here: \url{https://rise4fun.com/z3/tutorialcontent/sequences}.}
They support most of the constraints listed in Table \ref{tab:cons} of Sect. \ref{sec:prelim} apart from 
string reversal, characters count and 
context-free grammar language membership. However, it is worth noting that among the \scs approaches of 
Table \ref{tab:tools}, only Trau can handle 
context-free grammar language membership, and only G-Strings can deal with characters count.

G-Strings can also handle most of the constraints of Table \ref{tab:cons}---all of them except the 
context-free grammar constraints. Adding the support for a new string constraint is easier for a CP-based 
string solver because, unlike SMT-based solvers, this does not require the definition of a new decision procedure or an axiomatization based on previously defined theories. The only thing required is the definition of a new propagator for possibly pruning the domains of the variables involved in the constraint.

Summarizing, we can say that nowadays a modern ``general-purpose'' string solver must be able to handle string (in-)equality, length, concatenation, and regular expression membership constraints. With this in mind, closing the problem of the decidability of theory of word equations and arithmetic over length functions would be a significant breakthrough for string constraint solving, at least from the theoretical perspective.
In addition, handling constraints like $\indexof$, $\replace$, $\repall$ and string-number conversion is  highly recommended given their frequent use in modern software applications. 
Other constraints like, e.g., lexicographic ordering, string reverse or characters count are probably less used in practice but they can be surely useful for some specific applications.

\subsubsection{Performance}
We conclude Section \ref{sec:app} by discussing the performance of the \scs solvers in Table \ref{tab:tools}. Here we rely on 
what is reported in the corresponding papers. However, it is important to note that the reported evaluations might be influenced by factors such as the choice of the benchmarks, the encoding of the problems, the version and parameters configuration of the solvers. 
We report here the words of \cite{powerscs}: \textit{``[\scs approaches] mostly rely on hand crafted suites exhibiting that one tool is better than the others on this particular set of benchmark''}. 
It is outside the scope of this survey to report an empirical evaluation between different \scs approaches. 

Slog~\cite{slog} outperformed CVC4, Z3str2 and Norn on a benchmark of 20386 string analysis instances generated from real web applications via Stranger tool~\cite{stranger}. These instances contain regular expression operations such as union, concatenation, and replacement.
Curiously, no other approach from Table \ref{tab:tools} has been compared against Slog. In \cite{sloth} it is mentioned that \textit{``We have not experimented with other semi-decision procedures, such as [...] Slog, since they [...] often are not able to process input in the SMT-LIBv2 format, which would complicate the experiments.''} This is probably the reason why OSTRICH~\cite{ostrich} was evaluated on Slog benchmarks without including the Slog approach itself in the evaluation.

The approach in \oursolv is compared against Sloth and OSTRICH. The authors show that this approach is not better than Sloth and OSTRICH when it comes to solve \scs problems with constraints they both support. However, \oursolv can be more efficient for some particular classes of \scs problems where either Sloth or OSTRICH struggle (e.g., those involving the string reversal constraint).

OSTRICH~\cite{ostrich} extends and outperforms Sloth~\cite{sloth}. As Sloth, it can handle constraints (e.g., string transducers) that neither 
CVC4 nor Z3str3 can solve. However, in the Slog benchmarks without $\repall$ and in the Kaluza benchmarks CVC4 was more effective.

Qzy~\cite{qzy} compared against Norn but excluded CVC4 and Z3str2 \textit{``due to their lack of support
for negation of regular language membership''}. To the best of our knowledge,
already at that time both CVC4 and Z3str2 supported negative regular expressions. However, it is possible that the support for this constraint was still prototypical.

Z3-Trau~\cite{z3trau} outperformed CVC4, Z3seq and Z3str especially on string-number conversion 
benchmarks mainly collected with the Py-Conbyte~\cite{pyconbyte} concolic testing tool for Python. 
Some benchmarks also came from the symbolic execution of JavaScript programs. Curiously, Z3-Trau 
was not compared against its predecessor Trau~\cite{trau}.

In \cite{cvc4-ext1,cvc4-ext2,cvc4-cpoints} CVC4 was compared against Z3str2, Z3seq and OSTRICH on different benchmarks, often achieving the best results. Exceptions are, e.g., the Slog benchmark (where OSTRICH achieved the peak performance) or the unsatisfiable instances of the TermEq and Aplas benchmarks of \cite{cvc4-ext2} where Z3seq was better. The comparisons between CVC4 and Z3str* in these papers is  limited to Z3str2, because at that time Z3str3 was still unstable.

Nowadays, Z3 solver~\cite{z3-tool} offers two main ways to solve string constraints: the Z3str3 solver, using the theory of strings, and Z3seq, employing the theory of sequences. Z3seq is the default solver, while Z3-str and Z3str2 are now outdated. In \cite{z3str3RE} the authors show that Z3str3RE performs better than Z3seq overall, while in \cite{z3seq} Z3seq outperforms Z3str3. This is due to the fact that the benchmarks used are different, although other factors may have been affected the solvers' performance. 

A promising approach is Z3str4~\cite{z3str4}, a portfolio solver not currently part of Z3, which selects the \scs approach to run based on algorithm selection techniques. The Z3str4's website~\cite{z3str4-tool} reports an evaluation over 20 different benchmarks. Overall, Z3str4 was better than Z3str3 and Z3seq but slightly worse than CVC4.
The interesting thing here is that the results are not homogeneous along all the benchmarks. For example, on Kausler benchmark 
Z3str4 is far better than all the other approaches, while on BanditFuzz problems CVC4 dominates all the  others.

Norn~\cite{norn} and S3* variants~\cite{s3,s3p,s3hash} are probably a bit outdated nowadays. Norn is however still usable and publicly available at \cite{norn-tool}, although its last update dates back to 2015. At \cite{s3-tool} one can find the binaries of all the three versions of S3 string solver (viz. S3, S3P, and S3\#). 
The sources are available only for S3 and no longer updated since 2014.
However, it is noteworthy to quote the words of \cite{z3seq}: \textit{``Many of the advances previously developed in S3 are now integrated within Z3's default string solver''}.

Woorpje~\cite{woorpje} is a recent SAT-based \scs approach that showed promising results against 
CVC4, Z3str3, Z3seq, Norn and Sloth. It was evaluated on a benchmark consisting of five handcrafted
tracks containing randomly-generated word equations and length constraints. 
As the authors underline, these benchmarks are interesting because they challenge established solvers.
However, it would be also interesting to assess the performance of Woorpje over other well-established benchmarks (see Section \ref{sec:bench}).

G-Strings~\cite{dashed-string-jnl} is currently the state-of-the-art solver for CP problems 
with string variables and constraints. It improved $\gecodes$ with the dashed string abstraction, 
enabling a lazier unfolding and a higher-level reasoning.
Several empirical evaluations, on both handcrafted and existing benchmarks, showed its effectiveness: G-Strings was often able to outperform CVC4, Z3str3, Z3seq and Norn~\cite{dashed-string,sweep-based,lexfind,regular,repall,dashed-string-jnl}.
Unfortunately, outside the CP world no other \scs approach has been compared against G-Strings.
As we shall see in Section \ref{sec:tool}, the main reason is arguably the lack of a standard encoding
for \scs problems.

Only a few of the approaches not shown in the Table \ref{tab:tools} are still available or maintained.
The automata-based approaches MONA~\cite{mona-tool}, JSA~\cite{jsa-tool}, the PHP string analyzer~\cite{php-str-tool} described in \cite{php-str}, and Rex~\cite{rex-tool} are still available, but among them the only approach that looks still maintained is the MONA project.
Kaluza solver is still available at \cite{kaluza-tool} but no longer maintained. Analogously, 
$\gecodes$ is on-line~\cite{gecode_s-tool} but no longer developed.

\section{Technological aspects}
\label{sec:tool}
In this Section we consider the more practical aspects of string solving, by focusing in particular on the 
\scs modeling languages, benchmarks and applications that have been developed.

\subsection{Modeling}
\label{sec:mod}
As mentioned in Section \ref{sec:comp}, there is no \emph{lingua franca} for \scs problems even though a 
number of domain-specific languages have been proposed (e.g., \cite{fido,drex}).
SMT solvers share a common, standard language called \textit{SMT-LIB}~\cite{smtlib} to encode (also) \scs problems. Conversely, there is no standard modeling language for G-Strings and CP solvers in general, although \textit{MiniZinc}~\cite{minizinc} can be considered nowadays a de-facto standard.
This lack has certainly hampered the cross-comparisons between heterogeneous \scs approaches, hence  
making difficult a rigorous and comprehensive evaluation between them. 
% For example, DReX~\cite{drex} is a declarative language 
%expressing all the regular string-to-string transformations
%based on function \emph{combinators}~\cite{regcomb}. 
%The focus of \cite{drex} is on the evaluation the output of a DReX program on a given input string.
%In particular, the main contribution of \cite{drex} is the identification of a consistency restriction 
%on the use of combinators in DReX programs, and a single-pass evaluation algorithm for consistent programs. %with time complexity that is linear in the length of the input string and polynomial in the size of the program.

A good news for \scs modeling came in February 2020: based on an initial proposal by Nikolaj Bjørner, Vijay Ganesh, Raphaël Michel and Margus Veanes, a theory of Unicode-based strings and regular expressions became officially part of SMT-LIB standard.
This theory, which undoubtedly represents a milestone for \scs, includes the definition of string constants, variables, and functions such as, e.g., 
concatenation, length, lexicographic ordering, find/replace, regular expressions manipulation, converting from/to integers, and so on.\footnote{The full specification is available at \url{http://smtlib.cs.uiowa.edu/theories-UnicodeStrings.shtml}}

Let us see a practical example of a SMT-LIB 2.6 specification involving the theory of strings. Consider again the PHP code in Listing \ref{lst:ex}. An automated program analysis tool may want to prevent the occurrence of the  '\texttt{<}' character in the \texttt{\$www} variable. To do so, the tool may collect the constraints encountered during the (dynamic) symbolic execution of the program and then add an additional ``counterexample constraint''  imposing the occurrence of '\texttt{<}' in \texttt{\$www}. 
In this way, if the resulting problem is satisfiable we have found a witness telling us that 
the PHP program may be vulnerable.

Listing \ref{lst:smt1} shows a SMT-LIB 2.6 instance corresponding to the \scs problem described above.
Lines \ref{smt1:var0} and \ref{smt1:var1} define two string variables \texttt{www\_0} and \texttt{www\_1} used to model the PHP variable \texttt{\$www} respectively before and after the 
%call to \texttt{prog\_replace} 
statement \texttt{\$www = preg\_replace("/[\^{}A-Za-z0-9 .-@:\textbackslash/]/", "", \$www)}
at line \ref{ex:repl} of Listing \ref{lst:ex}.

Lines \ref{smt1:repl0}--\ref{smt1:repl1} capture the semantics of \texttt{prog\_replace} through the SMT-LIB 2.6 function \texttt{replace\_re\_all}, which replaces, left-to right, each shortest non-empty match of the given regular expression in \texttt{www\_0} by the empty string. 
The resulting string is then equalized to the \texttt{www\_1} variable.

The input pattern \texttt{[\^{}A-Za-z0-9 .-@:\textbackslash/]} of \texttt{prog\_replace}
is modelled by composing the regular expression functions \texttt{re.inter}, \texttt{re.allchar}, 
\texttt{re.comp}, \texttt{re.union}, \texttt{re.range}, and \texttt{str.to\_re}.
The resulting regular expression denotes the restricted alphabet
$\Sigma - \{\texttt{a}, \dots, \texttt{Z},\texttt{A}, \dots, \texttt{Z},0, \dots, 9,~,\texttt{.}, \dots, \texttt{@},\texttt{:},\texttt{/} \}$, where $\Sigma$ is the set of Unicode characters.

Finally, at line \ref{smt1:indexof} the \texttt{indexof} function is used to assert the presence of '\texttt{<}' in \texttt{www\_1} by imposing that its first occurrence starts at an index greater or equal than zero (note that string indexes are 0-based in SMT-LIB, so \texttt{indexof} returns $-1$ if the searched string does not occur in the target string).
\begin{figure*}[t]
\begin{lstlisting}[language=minizinc,escapechar=|,label=lst:smt1,caption=SMT-LIB 2.6 encoding of the \scs problem.]
(declare-fun www_0 () String) |\label{smt1:var0}|
(declare-fun www_1 () String) |\label{smt1:var1}|
(assert (= www_1 (str.replace_re_all www_0 (re.inter re.allchar (re.comp (re.union |\label{smt1:repl0}|
  (re.range "A" "Z") (re.range "a" "z") (re.range "0" "9") (str.to_re " ")
  (re.range "." "@") (str.to_re ":") (str.to_re "/"))) "" ))) |\label{smt1:repl1}|
(assert (>= (str.indexof www_1 "<") 0) 0) ) |\label{smt1:indexof}|
(check-sat)
\end{lstlisting}
\begin{lstlisting}[language=minizinc,label=lst:smt2,caption=Reformulation of Listing \ref{lst:smt1} without \texttt{str.replace\_re\_all}.]
(declare-fun www () String)
(assert (str.in_re www (re.* (re.union
  (re.range "A" "Z") (re.range "a" "z") (re.range "0" "9") (str.to_re " ") 
  (re.range "." "@") (str.to_re ":") (str.to_re "/")))))
(assert (>= (str.indexof www "<" 0) 0) )
(check-sat)
\end{lstlisting}
\begin{lstlisting}[escapechar=$,language=minizinc,caption=MiniZinc model translated from Listing \ref{lst:smt2} with \texttt{smt2mzn-str}.,label=lst:mzn]
var string: www;
constraint str_reg(www, "(([A-Z])|([a-z])|([0-9])|( )|([.-@])|(:)|(/))*"); $\label{mzn:regex}$
constraint ((str_find_offset("<", www, (0)+(1)))-(1)) >= (0); $\label{mzn:find}$
solve satisfy;
\end{lstlisting}
%\begin{lstlisting}[language=minizinc,label=ex:mzn2,caption=MiniZinc model of Listing \ref{ex:mzn1} simplified.]
%var string: www_0;
%constraint str_reg(www_0, "([A-Z]|[a-z]|[0-9]| |[.-@]|:|/)*");
%constraint str_find("<", www_0) >= 1;
%solve satisfy;
%\end{lstlisting}
\end{figure*}

State-of-the-art solvers like CVC4 and Z3 can parse the code in Listing \ref{lst:smt1}.
CVC4 can also instantaneously solve this problem, while Z3 cannot support the \texttt{replace\_re\_all} operator, a new entry in SMT-LIB.
This operator is hard to handle because, 
in addition to the already difficult $\repall$ constraint (see Table \ref{tab:cons} of Section \ref{sec:scs}), one has to handle also the search for a regular expression pattern in a target string.
However, in this particular case we can reformulate the problem as shown in Listing \ref{lst:smt2}.
The trick here is the transformation of \texttt{replace\_re\_all} into a logically equivalent 
regular membership constraint. In practice, instead of removing every string matching \texttt{[\^{}A-Za-z0-9 .-@:\textbackslash/]} from \texttt{www} we simply impose the membership of \texttt{www} in \texttt{[A-Za-z0-9 .-@:\textbackslash/]$^*$}.
With this encoding, also Z3 can instantaneously find a solution.

Let us now switch to the CP side. 
G. Gange developed \texttt{smt2mzn-str}~\cite{smt2mzn}, a prototypical compiler from SMT-LIB to MiniZinc able to process string variables and constraints. Unfortunately, the SMT-LIB syntax it supports is earlier than version 2.6. Having an up-to-date compiler is important to compare CP and SMT solvers. In particular, in this way the CP solvers could be submitted to the SMT competition.

Listing \ref{lst:mzn} shows how \texttt{smt2mzn-str} translates the SMT-LIB instance of Listing \ref{lst:smt2} to MiniZinc (Listing \ref{lst:smt1} cannot be translated because \texttt{replace\_re\_all}  
is not supported).
Line \ref{mzn:regex} of Listing \ref{lst:mzn} encodes the regular expression membership 
while line \ref{mzn:find} encodes the occurrence constraint. Note that here the name ``find'' is preferred to ``indexof''~\cite{lexfind} and that the string indexes are 1-based, so in general \texttt{str\_find\_offset(X,Y,K)} returns 0 if and only if \texttt{X} does not occur in \texttt{Y[K]$\cdots$Y[|Y|]}.

The MiniZinc model in Listing \ref{lst:mzn} enables a CP string solver like G-Strings to quickly solve 
the \scs problem. Unfortunately, this encoding is not standard because strings are not (yet) part of the 
official MiniZinc release, although the unofficial extension introduced in \cite{mzn-strings} 
is currently usable to model and solve string constraints. In fact, the G-Strings solver has an 
interface to process and solve MiniZinc models with strings.

\subsection{Benchmarks}
\label{sec:bench}
The use of compilers as a bridge between SMT-LIB and MiniZinc may greatly help the definition of standard benchmarks processable by both SMT and CP solvers. This may be a first step towards a closer and more fruitful interaction between these communities. For example, Bofill et al.~\cite{fzn2smt} defined a compiler from FlatZinc---the low-level language derived from MiniZinc---to SMT-LIB that was used to run the Yices SMT solver over the CP problems of the MiniZinc Challenges 2010--2012~\cite{mznc}.

Unsurprisingly, apart from some handcrafted MiniZinc models defined in \cite{sweep-based,lexfind,repall}, virtually all the string benchmarks that can be found in the literature are encoded in SMT-LIB language. 
A nice framework for the \scs benchmarks generation and evaluation is \emph{\strfuzz}~\cite{strfuzz}, a modular SMT-LIB problem instance transformer and generator for string solvers.
%It is a useful tool for string solver developers and testers, and can help expose bugs and performance issues.
A repository of several SMT-LIB 2.0/2.5 problem instances 
generated and transformed with \strfuzz is available online\footnote{http://stringfuzz.dmitryblotsky.com/benchmarks/}. 
Table \ref{tab:strfuzz}  
summarizes the nature of these instances, grouped into twelve different classes. 
\begin{table}[t]
	\centering
	\caption{\strfuzz benchmarks composition\label{tab:strfuzz}}
	% \scalebox{0.79}{
		\begin{tabular}{lll}
			\hline
			Class & Description & Quantity \\
			\hline
			\textit{Concats-\{Small,Big\}} & Right-heavy, deep tree of concats &  120 \\
			\textit{Concats-Balanced} & Balanced, deep tree of concats & 100 \\
			\textit{Concats-Extracts-\{Small,Big\}} & Single concat tree, with character extractions & 120 \\
			\textit{Lengths-\{Long,Short\}} & Single, large length constraint on a variable & 200 \\
			\textit{Lengths-Concats} & Tree of fixed-length concats of variables & 100 \\
			\textit{Overlaps-\{Small,Big\}} & Word equations $aX = Xb$ with $X$ variable and $a,b$ literals& 80 \\
			\textit{Regex-\{Small,Big\}} & Complex regex membership test & 120 \\
			\textit{Many-Regexes} & Multiple random regex membership tests & 40 \\
			\textit{Regex-Deep} & Regex membership test with many nested operators & 45 \\
			\textit{Regex-Pair} & Test for membership in one regex, but not another& 40 \\
			\textit{Regex-Lengths} & Regex membership test, and a length constraint & 40 \\
			\textit{Different-Prefix} & Equality of two deep concats with different prefixes & 60 \\
			\hline
	\end{tabular}%}
\end{table}

\strfuzz also reports the performance of Z3str3, CVC4, Z3seq, and Norn on such instances. Only these SMT solvers were used because other \scs approaches were either unstable or could not properly process the proposed SMT-LIB syntax. In \cite{dashed-string-jnl}, G-Strings has been compared against, and often outperformed, the solvers above after translating the generated \strfuzz instances into MiniZinc with the \texttt{smt2mzn-str} compiler.

A more recent fuzzer for SMT solvers is \emph{BanditFuzz}~\cite{banditfuzz}. BanditFuzz uses the  
abstract syntax tree generation procedure of \strfuzz for generating random SMT instances, and also extends it for handling floating point arithmetic. The key feature of BanditFuzz is the isolation of the cause for a performance issue, enoded in the form of grammatical constructs such as, e.g., predicates or functions. To learn which constructs are most likely to cause performance issues, 
\textit{reinforcement learning} methods are used---in particular the problem of how to optimally mutate an input is reduced to the multi-arm bandit problem~\cite{rl}.

SMT-LIB string benchmarks are also available on the website of the SMT competition~\cite{smtcomp}, that since 2018 has also a single-query string track. In 2020 for the first time two different solvers (viz. Z3str4 and CVC4, with the latter achieving better results) have entered the competition, while in 2018 and 2019 editions only CVC4 participated.

A relevant set of SMT-LIB benchmarks have been collected in \cite{powerscs}. This set includes the following benchmarks:
\begin{itemize}
\item \textit{Kaluza}: 47284 instances generated from the dynamic symbolic execution of JavaScript by the Kudzu tool~\cite{kaluza}
\item \textit{PyEx}: 8414 instances generated by PyEx~\cite{pyex}, a symbolic executor for Python programs, by Reynolds et al~\cite{cvc4-ext1}
\item \textit{PISA}: 12 instances derived from real-world Java sanitizer methods~\cite{pisa}
\item \textit{AppScan}: 8 instances collected by using the output of security warnings generated by IBM Security AppScan~\cite{z3str}
\item \textit{\strfuzz}: 1065 instances generated with \strfuzz tool~\cite{strfuzz}
\item \textit{Norn}: 1027 instances coming from queries generated during verification of string-processing programs~\cite{norn}
\item \textit{LightTrau}: 100 instances generated by Abdulla et al.~\cite{trau} for testing unsatisfiable formulae
\item \textit{Woorpje}: 809 instances handcrafted by Day et al.~\cite{woorpje} to evaluate Woorpje tool
\item \textit{Joaco}: 94 instances based on 11 open-source Java Web applications and security benchmarks~\cite{acosolver}
\item \textit{Kausler}: 120 instances derived from 8 Java programs via dynamic symbolic execution by Kausler et al.~\cite{str-eval}
\item \textit{Cashew}: 394 instances from Kaluza benchmark set normalised via Cashew tool~\cite{cashew}
\item \textit{Stranger}: 4 instances manually translated from 4 PHP real-world web applications.
\end{itemize}
The benchmarks' collection is not the only contribution of \cite{powerscs}, which also introduces the \zalig framework to facilitate the performance evaluation and comparison of SMT-based string solvers over those benchmarks.

In addition to those of \zalig, the Z3str4 website~\cite{z3str4-tool} offers other benchmarks such as
Leetcode Strings,
Sloth,
Z3str3 Regression,
BanditFuzz. Unfortunately, there is no description of them.

\subsection{Applications}
The main application field for string solving is undoubtedly the
area of software verification and testing.
In fact, most of the \scs approaches presented in Section \ref{sec:app} were developed to facilitate 
the automated detection of web vulnerabilities.

Kausler et al.~\cite{str-eval} 
performed an evaluation of string constraint solvers in the context of \textit{symbolic execution}~\cite{king1976}.
What they found out is that, as one can expect, one solver might be more appropriate than another
depending on the input program. This is also pointed out in \cite{aratha}, where Amadini et al. 
presented a multi-solver tool for the dynamic symbolic execution of JavaScript
built on the top of CVC4, G-Strings and Z3 solvers. These observations probably underlie the development of Z3str4~\cite{z3str4}.

In \cite{DBLP:conf/sigsoft/BangAPPB16}, a combined approach 
based on symbolic execution, string analysis and
model counting is used to detect and quantify side-channels vulnerabilities in Java programs. 
This framework is built on the top of Z3 solver~\cite{z3} and ABC~\cite{abc} model counter.

In \cite{PHPRepair}, the PHPRepair tool is used for automatically repairing HTML generation errors in 
PHP via string constraint solving. The property that all tests of a suite should 
produce their expected output is modelled with string constraints encoded into the language of Kodkod~\cite{kodkod}, a SAT-based constraint solver.

ACO-Solver~\cite{acosolver} is a tool for the detection of injections and XSS vulnerabilities 
for Java Web applications that uses a hybrid procedure based on the \emph{ant colony optimization} meta-heuristic. ACO-Solver is a meta-solver that complements the support for string operations provided by existing string solvers. JOACO-CS is an extension of ACO-Solver publicly available at \cite{joaco-tool}.

Abstracting a set of strings with a finite formalism is not merely a 
string solving affair. For example, the well-known \emph{Abstract Interpretation}~\cite{CousotC77} framework may require the sound approximation of sets of strings---i.e., all the possible ``concrete'' values that a string variable of the input program can take---with an abstract counterpart. 

Several abstract domains have been proposed to approximate 
sets of strings. These domains vary according to the properties that 
one needs to capture (e.g., the string length, the prefix or suffix, 
the characters occurring in a string).
Madsen et al.~\cite{DBLP:conf/cc/MadsenA14} proposed and evaluated a suite of twelve string domains for the static analysis of dynamic field access.
Additional string domains are also discussed in \cite{DBLP:journals/spe/CostantiniFC15}.

In \cite{m-string}, a refined segmentation abstract domain called M-String is introduced for the static analysis of strings in the C programming language. M-String is based the parametric segmentation domain introduced by P. Cousot for the representation of arrays.

Choi et al.~\cite{DBLP:conf/aplas/ChoiLKD06} used restricted regular expressions as an abstract
domain for strings in the context of Java analysis. Park et al.~\cite{DBLP:conf/dls/ParkIR16} use a stricter variant of this idea, with a more clearly defined string abstract domain.

Amadini et al.~\cite{js-string} provided an evaluation on the combination via direct product of different string abstract domains in order to improve the precision of JavaScript static analysis.
In \cite{ref-domains} this combination is achieved
via reduced product by using the set of regular languages as a \emph{reference domain} for the other string domains.

Finally, we mention that string solving techniques might be useful 
in the context of \emph{Bioinformatics}, where a number of CP techniques has been already applied (see, e.g., the works by Barahona et al.~\cite{DBLP:journals/constraints/BarahonaK08,DBLP:conf/aime/KrippahlMB13,DBLP:conf/cp/KrippahlB16}).
\section{Conclusions}
\label{sec:concl}
In this work we provided a comprehensive survey on the various aspects of string constraint solving (\scs), an emerging important field orthogonal to combinatorics on words and constraint solving. 
We focused in particular on the technological aspects of string solving, by grouping the main \scs approaches we are aware, from the early 
proposals to the state-of-the-art approaches, into three main categories: automata-based, word-based and unfolding-based.

After exploring several approaches of the \scs galaxy, we can safely say that the main application 
for string solving is the field of software testing and verification, with particular emphasis on the detection of vulnerabilities conveyed by improper use of strings.
Another conclusion we can draw is that ``general-purpose'' \scs approaches nowadays should probably take advantage of the advances that SAT/SMT and CP technologies made in this field over the last years to 
handle the string constraints most frequently occurring in modern programs.

Among the future directions for string constraint solving we mention the four challenges reported in \cite{ecai-str}, namely:
\begin{itemize}
	\item \emph{Extend} the \scs capabilities to properly handle complex string operations, frequently occurring in web programming~\cite{expose-regex}, such as \emph{back-references}, \emph{lookaheads/lookbehinds} or \emph{greedy matching}. These operations significantly extend the expressiveness of regular expressions in a way that string solvers may have difficulty with---e.g., it is known that regular expressions with backreferences can describe non-regular languages.
	\item \emph{Improve} the efficiency of \scs solvers with new algorithms and search heuristics. At present, some empirical evidences (e.g. \cite{dashed-string,sweep-based,lexfind}) show that SMT solvers tend to fail when dealing with string literals, while CP strings solvers may struggle to prove unsatisfiability because they do not have (yet) a support for \textit{clause learning}---probably the key to success  for CP solvers over integers like Chuffed~\cite{chuffed} or OR-Tools~\cite{or-tools}.
	\item \emph{Combine} \scs solvers with a \emph{portfolio approach}~\cite{kotthoff2016algorithm} in order to exploit their different nature and uneven performance across different problem instances.
	Recent experiments for the dynamic symbolic execution of JavaScript already show some potential for this approach, that can be seen as an instance of the \textit{algorithm selection} problem~\cite{kotthoff2016algorithm}. The novel Z3str4 solver~\cite{z3str4-tool} also includes an algorithm selection architecture choosing the best algorithms to run based on their expected performance.
	\item \emph{Use} \scs solvers and related tools in different fields. The best candidates are probably software verification and testing, model checking and cybersecurity. In these areas \scs plays a crucial role, given the massive use of string operations in modern (web) applications.
\end{itemize}

Finally, we hope that the research in \scs will encourage a closer and more fruitful collaboration between the CP and the SAT/SMT communities. A step in this direction was taken by 
Bardin et al. in a 2019 Dagstuhl seminar~\cite{bardin_et_al:DR:2019:10857}.

\begin{acks}
We would like to thank the anonymous reviewers for their informative and constructive comments. 
Many thanks also to Murphy Berzish, Pierre Flener, Vijay Ganesh, Andrew Reynolds, Peter J. Stuckey, and Cesare Tinelli for their help and feedback.
\end{acks}

%%
%% The next two lines define the bibliography style to be used, and
%% the bibliography file.
\bibliographystyle{ACM-Reference-Format}
\bibliography{biblio}

\end{document}